\documentclass[acmtog]{acmart}

\AtBeginDocument{%
  }

\definecolor{Coral}{rgb}{1, 0.47, 0.24}

\newcommand{\themethod}{AnySplat\xspace}

\usepackage{booktabs}
\usepackage{multirow}
\usepackage{graphicx}
\usepackage{array}
\usepackage{gensymb}  % Provides the \degree command
\usepackage{fancyvrb}
\usepackage{lipsum}
\usepackage{xspace}
\usepackage{pifont}

\acmSubmissionID{1712}

\begin{document}

\setcopyright{acmlicensed}
\acmJournal{TOG}
\acmYear{2025} \acmVolume{44} \acmNumber{6} \acmArticle{} \acmMonth{12}\acmDOI{10.1145/3763326}

\title{AnySplat: Feed-forward 3D Gaussian Splatting from Unconstrained Views}

\author{Lihan Jiang}
\orcid{0009-0001-2899-273X}
\authornote{Equal Contribution, Alphabetical Order.}
\affiliation{
  \institution{University of Science and Technology of China}
  \city{Anhui}
  \country{China}
}
\affiliation{
  \institution{Shanghai Artificial Intelligence Laboratory}
  \city{Shanghai}
  \country{China}
}
\email{mr.lhjiang@gmail.com}

\author{Yucheng Mao}
\orcid{0009-0007-7124-4431}
\authornotemark[1]
\affiliation{
  \institution{Shanghai Artificial Intelligence Laboratory}
  \city{Shanghai}
  \country{China}
}
\email{yucheng.mao.cs@gmail.com}

\author{Linning Xu}
\orcid{0000-0003-1026-2410}
\affiliation{
    \institution{The Chinese University of Hong Kong}
    \city{Hong Kong}
    \country{China}
}
\email{linningxu@link.cuhk.edu.hk}

\author{Tao Lu}
\orcid{0009-0000-8830-3820}
\affiliation{
    \institution{Brown University}
    \city{Rhode Island}
    \country{United States of America}
}
\email{tao_lu@brown.edu}

\author{Kerui Ren}
\orcid{0009-0003-8010-5733}
\affiliation{
    \institution{Shanghai Jiao Tong University}
    \city{Shanghai}
    \country{China}
}
\affiliation{
  \institution{Shanghai Artificial Intelligence Laboratory}
  \city{Shanghai}
  \country{China}
}
\email{renkerui@sjtu.edu.cn}

\author{Yichen Jin}
\orcid{0009-0005-6213-2038}
\affiliation{
  \institution{Shanghai Artificial Intelligence Laboratory}
  \city{Shanghai}
  \country{China}
}
\email{13905152060@163.com}

\author{Xudong Xu}
\orcid{0009-0003-8858-0918}
\affiliation{
  \institution{Shanghai Artificial Intelligence Laboratory}
  \city{Shanghai}
  \country{China}
}
\email{xuxudong@pjlab.org.cn}

\author{Mulin Yu}
\orcid{0000-0002-0327-4547}
\affiliation{
  \institution{Shanghai Artificial Intelligence Laboratory}
  \city{Shanghai}
  \country{China}
}
\email{yumulin@pjlab.org.cn}

\author{Jiangmiao Pang}
\orcid{0000-0002-6711-9319}
\affiliation{
  \institution{Shanghai Artificial Intelligence Laboratory}
  \city{Shanghai}
  \country{China}
}
\email{pangjiangmiao@gmail.com}

\author{Feng Zhao}
\orcid{0000-0001-6767-8105}
\affiliation{
  \institution{University of Science and Technology of China}
  \city{Anhui}
  \country{China}
}
\authornote{Corresponding Author}
\email{fzhao956@ustc.edu.cn}

\author{Dahua Lin}
\orcid{0000-0002-8865-7896}
\affiliation{
    \institution{The Chinese University of Hong Kong}
    \city{Hong Kong}
    \country{China}
}
\email{dhlin@ie.cuhk.edu.hk}

\author{Bo Dai}
\orcid{0000-0003-0777-9232}
\affiliation{
  \institution{The University of Hong Kong}
  \city{Hong Kong}
  \country{China}
}
\authornotemark[2]
\email{bdai@hku.hk}

\renewcommand{\shortauthors}{Jiang et al.}

\begin{teaserfigure}
    \centering
    \includegraphics[width=\linewidth]{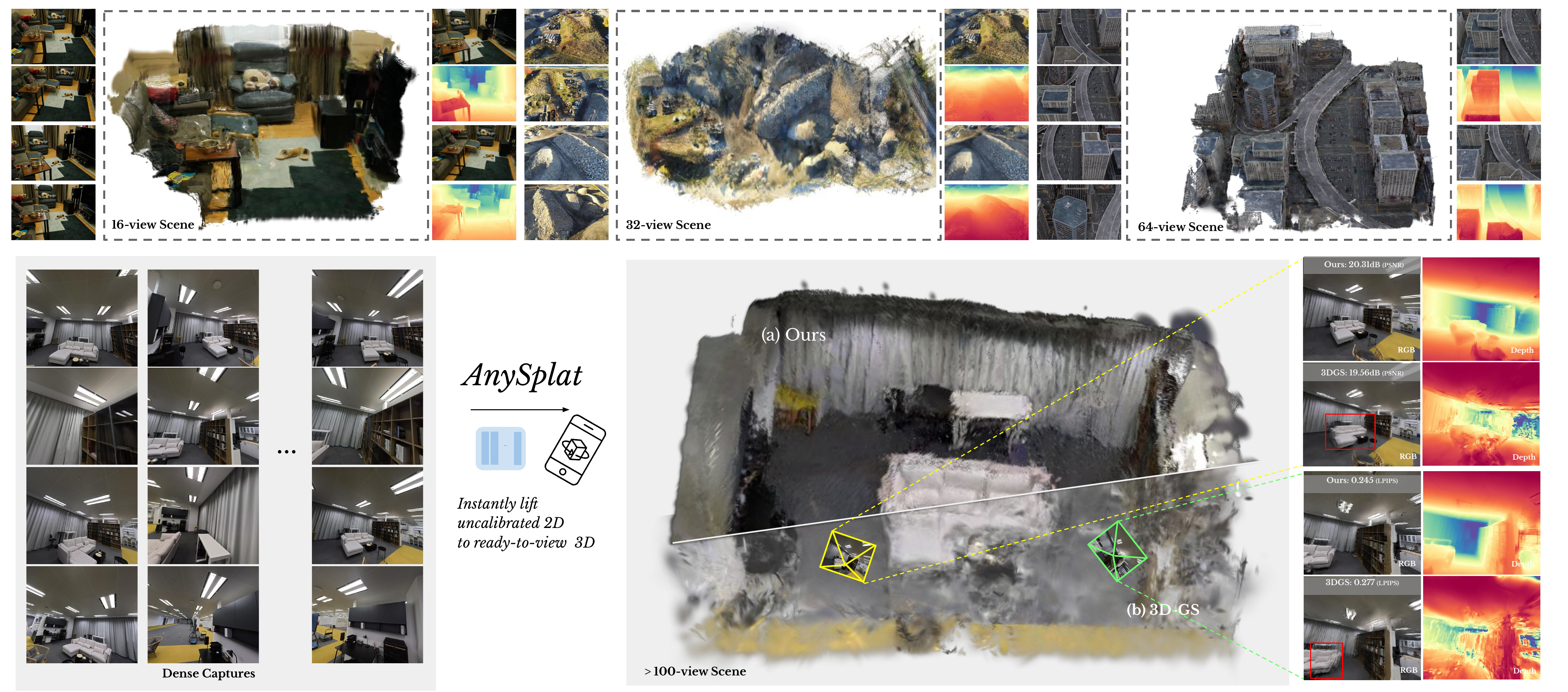}
    \caption{\label{fig:teaser}
        AnySplat lifts multi-view captures, from sparse to dense, into ready-to-view 3D scenes represented with 3D Gaussians~\cite{kerbl20233d}. Unlike previous multi-view reconstruction and neural rendering methods, which rely on precise camera calibration, tedious per-scene optimization, and are often sensitive to input noise, AnySplat robustly handles a wide variety of capture scenarios in just seconds.
    }
    \Description{teaser}
\end{teaserfigure}

\begin{abstract}
We introduce \themethod, a feed‑forward network for novel‑view synthesis from uncalibrated image collections. In contrast to traditional neural‑rendering pipelines that demand known camera poses and per‑scene optimization, or recent feed‑forward methods that buckle under the computational weight of dense views—our model predicts everything in one shot. A single forward pass yields a set of 3D Gaussian primitives encoding both scene geometry and appearance, and the corresponding camera intrinsics and extrinsics for each input image. This unified design scales effortlessly to casually captured, multi‑view datasets without any pose annotations. In extensive zero‑shot evaluations, \themethod matches the quality of pose‑aware baselines in both sparse‑ and dense‑view scenarios while surpassing existing pose‑free approaches. Moreover, it greatly reduces rendering latency compared to optimization‑based neural fields, bringing real‑time novel‑view synthesis within reach for unconstrained capture settings. 
Project page: 
\href{https://city-super.github.io/anysplat/}{\textcolor{magenta}{\textbf{https://city-super.github.io/anysplat/}}}.

\end{abstract}





\maketitle

\section{Introduction}

Recent advances in 3D foundation models \cite{wang2024dust3r,yang2025fast3r,wang2025vggt} have reshaped how we view the problem of reconstructing 3D scenes from 2D images. By inferring dense point clouds from a single view to thousands within seconds, these methods streamline or even eliminate traditional multi-stage reconstruction pipelines, making 3D scene reconstruction more accessible across a wider range of applications.

Despite their powerful geometry priors, current foundation models often struggle to capture fine detail, photorealism, and geometric consistency—especially when processing highly overlapping inputs, which can yield misaligned or noisy reconstructions. By contrast, novel-view synthesis (NVS) methods such as NeRF \cite{mildenhall2021nerf} and its recent extensions \cite{kerbl20233d} deliver exceptional rendering fidelity, but only by offloading the hard work to a costly preprocessing stage. These pipelines first estimate camera poses via structure-from-motion and then perform per-scene neural field optimization.
This delay between capture and usable output, along with computation costs that grow with the number of input frames, limits their practical applicability in many real-world scenarios.

Witnessing this paradigm shift brought by feed-forward architectures like ViT~\cite{dosovitskiy2020image} in 3D modeling, we ask: can novel-view synthesis (NVS) from multiview captures naturally benefit? To bridge the gap between geometry priors and ``ready-to-see'' output, as exemplified in Fig.~\ref{fig:teaser}, we augment the foundation model with a lightweight rendering head. During training, this head refines and synthesizes appearance via a pseudo-label distillation training strategy, no ground-truth 3D annotations required, thereby injecting texture priors and enforcing geometric coherence in a single, end-to-end pass. 
This training strategy paves the way for extending the reach of 3D foundation models \cite{wang2024dust3r,yang2025fast3r,wang2025vggt} far beyond finite, annotated datasets—enabling seamless generalization to unbounded new scenes with minimal overhead.

Specifically, we propose \themethod, a feed-forward network for novel view synthesis trained on unconstrained and unposed multi-view images. \themethod employs a geometry transformer to encode these images into high-dimensional features, which are then decoded into Gaussian parameters and camera poses. To improve efficiency, we introduce a differentiable voxelization module that merges pixel-wise Gaussian primitives into voxel-wise Gaussians, eliminating 30–70\% of redundant primitives while maintaining comparable rendering quality.
Since 3D annotations in real-world scenarios are often noisy, we design a novel pseudo-label knowledge distillation pipeline. In this framework, we distill camera and geometry priors from pretrained VGGT~\cite{wang2025vggt} backbone as external supervision. As a result, \themethod can be trained without any 3D SfM or MVS supervision, relying solely on uncalibrated images, making it promising to scale up to unconstrained capture with readily usable input.
We train \themethod on nine diverse and large-scale datasets, exposing the model to a wide range of geometric and appearance variations. As a result, our method demonstrates superior zero-shot generalization performance on unseen datasets. Experimental results show that \themethod achieves excellent novel view synthesis quality, more consistent geometry, more accurate pose estimation, and faster inference times compared to both state-of-the-art feed-forward and optimization-based methods.

In summary, our key contributions are:
\begin{itemize} 
\item \emph{Feed-forward reconstruction and rendering}. Our model takes uncalibrated multi-view inputs and simultaneously predicts 3D Gaussian primitives and their camera intrinsics/extrinsics, delivering higher‐quality reconstructions than prior feed-forward methods—and even outperforming optimization-based pipelines in challenging scenarios.

\item \emph{Efficient pseudo-label knowledge distillation.} We distill geometry and texture priors from a pretrained VGGT model via a novel, end-to-end training pipeline—with only RGB images—unlocking high-fidelity rendering and enhanced multi-view consistency in under one day on 8–16 GPUs.

\item \emph{Differentiable voxel-guided Gaussian pruning. }
Our custom voxelization strategy eliminates 30–70 \% of Gaussian primitives while preserving rendering quality, yielding a unified, compute‐efficient model that gracefully handles both sparse and dense capture setups.

\end{itemize}

\section{Related Work}

\subsection{Optimization-based Novel View Synthesis Methods.}
Neural Radiance Fields (NeRF)~\cite{mildenhall2021nerf, barron2022mip, muller2022instantngp} pioneered high-quality novel view synthesis by learning continuous volumetric density and radiance fields via coordinate-based networks, but its reliance on expensive volume rendering precludes real-time performance. 
In contrast, 3D Gaussian Splatting (3DGS)~\cite{kerbl20233d, yu2024mip, scaffoldgs, ren2024octree, jiang2024horizon, yang2025virtualized, yu2024gsdf, feng2025flashgs, lu2024turbo} explicitly represents scenes with millions of anisotropic Gaussians and exploits differentiable rasterization to render photorealistic views at over 30 FPS (1080p). Its core advances—adaptive density control for geometry refinement and spherical harmonics for view-dependent shading—enable real-time playback.
Despite these advances, most NeRF and 3DGS methods assume access to accurate camera poses, typically obtained via classical Structure-from-Motion tools such as COLMAP~\cite{schonberger2016structure} or other relevant methods~\cite{brachmann2024scene, wang2024vggsfm, pan2024global}. This requirement introduces an implicit preprocessing step that conceals the significant time and logistical costs of large-scale, multi-view data acquisition and registration.
To address these limitations, recent approaches attempt to jointly optimize camera poses and scene representation. However, they either require incremental image sequences and intrinsics~\cite{monogs,keetha2024splatam,yan2024gs,fu2024colmap, meuleman2025onthefly} as input or are limited to scenarios with minimal motion~\cite{meng2021gnerf,wang2021nerf} or sparse view coverage~\cite{fan2024instantsplat}. Furthermore, these methods still involve redundant optimization processes. In contrast, \themethod can directly predict 3D Gaussians and camera parameters within seconds, significantly accelerating the 3D reconstruction process.

\subsection{Generalizable 3D Reconstruction Methods.}
Most view synthesis methods require tens of minutes or even hours to optimize on densely captured data. Recently, several generalizable 3D reconstruction methods have been proposed, 
which can be broadly categorized into two types: pose-aware methods, which assume known camera parameters, and pose-free methods, which jointly infer both geometry and camera poses.
% pose-aware and pose-free methods. 

\paragraph{Pose-aware generalizable methods.} Pose-aware generalizable methods rapidly reconstructed 3D models from calibrated images and their corresponding poses. 
These approaches can be broadly categorized into three methodological strands: (1) 3D Gaussian Splatting based techniques~\cite{charatan2024pixelsplat, chen2024mvsplat, chen2024mvsplat360, wang2024freesplat, xu2024depthsplat} which directly predict 3D Gaussian primitive as the scene representation, (2) neural network based frameworks~\cite{yu2021pixelnerf, wang2021ibrnet, chen2021mvsnerf, flynn2024quark, jin2024lvsm, jiang2025rayzer} employing neural network to infer the appearance of the novel view image without any 3D representation, and (3) the emerging LRM architecture family~\cite{lrm,gslrm, xu2024grm, ziwen2024long}. Despite these pose-aware reconstruction methods significantly reducing optimization time and improving performance under sparse-view conditions, their broader applicability remains limited due to the necessity for accurate image poses as input.
\paragraph{Pose-free generalizable methods.} To achieve truly end-to-end 3D reconstruction, pose-free generalizable methods rely solely on images as input, and most of them simultaneously predict image poses alongside the reconstructed 3D model. Among them, Dust3R~\cite{wang2024dust3r} and extended by MASt3R~\cite{mast3r}, replace traditional multi-stage pipelines with a single large-scale model that jointly predicts depth and fuses it into a dense scene. More recent methods~\cite{wang20243d, liu2025slam3r, murai2025mast3r, wang2025continuous, wang2025vggt, yang2025fast3r, tang2025mv}, cascade transformer blocks to jointly infer camera poses, point trajectories, and scene geometry in a single forward pass, achieving substantial improvements in both accuracy and runtime. While these methods highlight the potential for efficiently scaling up 3D asset reconstruction, they generally struggle in poor texture representation and multi-view misalignment problem, which significantly hinder their novel view synthesis performance. Another line of work~\cite{jiang2023leap, wang2023pf, hong2024pf3plat, ye2024no,zhang2025flare,smart2024splatt3r,chen2024pref3r} targets novel view synthesis from unposed images, but these methods only work in sparse-view settings.
\section{Method}

\begin{figure*}[t!]
    \centering
    \includegraphics[width=\linewidth]{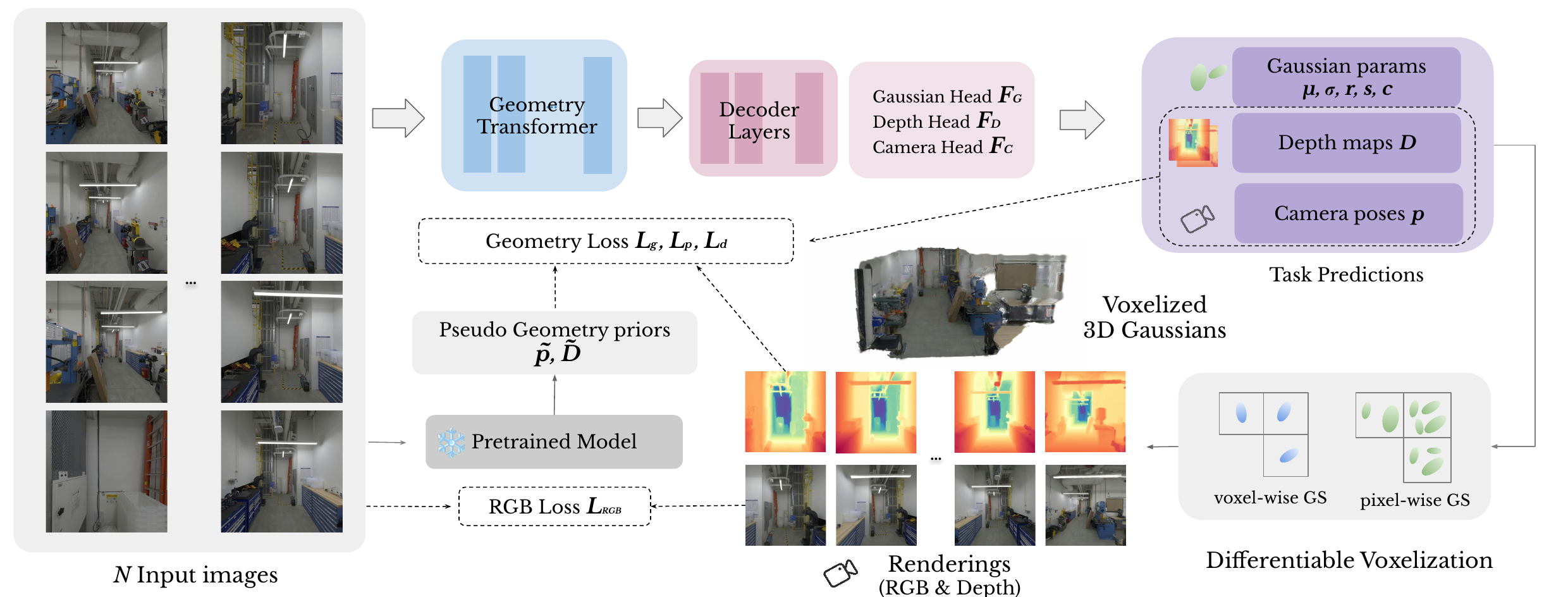}
    \caption{\textbf{Overview of \themethod.} Starting from a set of uncalibrated images, a transformer-based geometry encoder is followed by three decoder heads: $\mathrm{F}_G$, $\mathrm{F}_D$, and $\mathrm{F}_C$, which respectively predict the Gaussian parameters ($\boldsymbol{\mu}, \sigma, \boldsymbol{r}, \boldsymbol{s}, \boldsymbol{c}$), the depth map $D$, and the camera poses $p$. 
 These outputs are used to construct a set of pixel-wise 3D Gaussians, which is then voxelized into pre-voxel 3D Gaussians with the proposed Differentiable Voxelization module. From the voxelized 3D Gaussians, multi-view images and depth maps are subsequently rendered. The rendered images are supervised using an RGB loss against the ground truth image, while the rendered depth maps, along with the decoded depth $D$ and camera poses $p$, are used to compute geometry losses. The geometries are supervised by pseudo-geometry priors ($\tilde{D}, \tilde{p}$) obtained by the pretrained VGGT~\cite{wang2025vggt}.
}
    \label{fig: pipeline}
\end{figure*}

% \subsection{Problem Set-ups}
% Our model only takes $N$ \textbf{uncalibrated} multi-view images $(I_i \in \mathbb{R}^{H \times W \times 3})_{i=1}^N$ as input, which are observed in the same 3D scene. \themethod maps the input images into a set of 3D Gaussians in the 3D space that represents the geometry and the appearance of the observed 3D scene; also, our model could jointly estimate the camera parameters $\ p_i \in \mathbb{R}^9$ (intrinsics and extrinsics) of image $I_i$:
% \begin{equation}
% f_{\boldsymbol{\theta}}:\left\{\boldsymbol{I}^n\right\}_{n=1}^N \mapsto\left\{ \cup_{g=1}^G \left(\boldsymbol{\mu}_g, \sigma_g, \boldsymbol{r}_g, \boldsymbol{s}_g, \boldsymbol{c}_g\right)\right\} \cup \left\{(p_i)_{i=1}^N\right\}, 
% \label{eq:mapping}
% \end{equation}

% We introduce \themethod, a large neural network that can take a various number (2-200) of uncalibrated images as input and produces 3D gaussian primitives as the representation of reconstructed 3D scene. We start by introducing our problem setup in Sec~\ref{sec:problem}, the dataset used during our training in Sec~\ref{sec:train_dataset}, model pipeline in Sec~\ref{sec:pipeline} and the training and inference strategy in Sec~\ref{sec:strategy}

We propose \themethod, a transformer-based neural network designed for rapid 3D scene reconstruction tailored for novel-view synthesis. Given uncalibrated images, from a single view up to hundreds, \emph{\themethod} directly predicts a set of 3D Gaussian primitives that compactly represent the reconstructed scene.

In the following sections, we first formalize our problem setup in Sec. \ref{sec:problem}, 
% then describe the training dataset in Sec. \ref{sec:train_dataset}, 
detail the model’s architecture and pipeline in Sec. \ref{sec:pipeline}, and finally present our training and inference strategies in Sec. \ref{sec:strategy}.

\subsection{Problem Setup}
\label{sec:problem}
% \uml{There is no background described in this section.}
Consider $N$ \emph{uncalibrated} views of a single 3D scene, given as images
$\{I_i\}_{i=1}^N$, where $I_i \in \mathbb{R}^{H\times W\times 3}$, \themethod aims to jointly reconstruct the scene geometry and appearance by predicting a) a collection of $G$ anisotropic 3D Gaussians
\begin{equation}
    \bigl\{(\boldsymbol{\mu}_g, \sigma_g, \boldsymbol{r}_g, \boldsymbol{s}_g, \boldsymbol{c}_g)\bigr\}_{g=1}^G,
\end{equation}
where each Gaussian is parameterized by a center position \(\boldsymbol{\mu}\in\mathbb{R}^3\), a positive opacity \(\sigma\in\mathbb{R}^+\), an orientation quaternion \(\boldsymbol{r}\in\mathbb{R}^4\), an anisotropic scale \(\boldsymbol{s}\in\mathbb{R}^3\), and a color embedding \(\boldsymbol{c}\in\mathbb{R}^{3\times (k+1)^2}\) represented via spherical‑harmonic coefficients of degree \(k\), following practice of ~\cite{kerbl20233d};
% its center $\boldsymbol{\mu}_g$, opacity $\sigma_g$, rotation $\boldsymbol{r}_g$, scale $\boldsymbol{s}_g$, 
% % scale $\sigma_g$, anisotropy axes $\boldsymbol{r}_g$, anisotropy extents $\boldsymbol{s}_g$, 
% and color $\boldsymbol{c}_g$; 
and 2) the camera parameters for each view 
\begin{equation}
    \{p_i\in\mathbb{R}^9\}_{i=1}^N,
\end{equation}
with $p_i$ encoding the intrinsics and extrinsics of image~$I_i$.
Formally, our model implements the mapping:
\begin{equation}
    f_{\boldsymbol{\theta}}\!:\;\{I_i\}_{i=1}^N
    \;\longmapsto\;
    \Bigl\{
      (\boldsymbol{\mu}_g, \sigma_g, \boldsymbol{r}_g, \boldsymbol{s}_g, \boldsymbol{c}_g)
    \Bigr\}_{g=1}^G
    \;\cup\;
    \{p_i\}_{i=1}^N.
\label{eq:mapping}
\end{equation}

\noindent We evaluate our model on two core tasks: novel view synthesis and multi-view camera pose estimation. 
Notably, this pipeline also produces several useful by-products—such as a global point map, per-frame depth maps, and associated confidence scores—that can support a variety of downstream applications.
% Supervising these by-products in the training process directly enhances our model's preformance and \yum{TBD Here}

\subsection{Pipeline}
\label{sec:pipeline}

Fig.~\ref{fig: pipeline} illustrates the overall pipeline of framework. In a nutshell, our model begins by encoding a set of uncalibrated multi-view images into high-dimensional feature representations, which are then decoded into both 3D Gaussian parameters and their corresponding camera poses. 
% To manage the explosion of Gaussians under dense views, 
To manage the linear growth in per-pixel Gaussians under dense views,
we introduce a differentiable voxelization module that clusters primitives into voxels, significantly reducing computational cost and facilitating smoother gradient flow. 

\paragraph{Geometry Transformer}  
Following VGGT~\cite{wang2025vggt}, we begin by patchifying each image \(I_i\) into \(l_I = \tfrac{H\,W}{p^2}\) tokens of dimension \(d\) using DINOv2~\cite{oquab2023dinov2}, where \(p=14\) and \(d=1024\).  To each image’s token sequence \(t_i^I\in\mathbb R^{l_I\times d}\), we prepend a learnable camera token \(t_i^g\in\mathbb R^{1\times d}\) and four register tokens \(t_i^R\in\mathbb R^{4\times d}\); for the first view only, we omit positional encodings on these tokens.  The combined tokens \(\bigl[t_i^I; t_i^g; t_i^R\bigr]\) from all \(N\) views are processed by an \(L\)-layer Alternating‑Attention transformer: each layer applies a frame attention over tokens of shape \(\mathbb R^{N\times(l_I+5)\times d}\), then a global attention over all views jointly as \(\mathbb R^{1\times N(l_I+5)\times d}\).

\paragraph{Camera Pose Prediction}
Camera pose estimation is essential for geometry reconstruction via novel-view rendering. The refined camera tokens \(\hat t_i^g\) are passed through the camera decoder $F_C$, which consists of four additional self-attention layers followed by a linear projection head, to predict each camera parameters \(p_i\). As in prior work, we set the first camera pose to the identity transformation and express all remaining poses in that shared local coordinate frame.
% The refined camera tokens \(\hat t_i^g\) pass through 4 additional self‑attention layers and a linear head to regress the camera parameters \(p_i\). Following previous works, the camera pose for the first camera is set to the identity, and the remaining camera poses are defined in the same local coordinate system.

\paragraph{Pixel‑wise Gaussian Parameter Prediction}  
As shown in Fig.~\ref{fig: pipeline}, we adopt a dual‑head design based on the DPT decoder~\cite{ranftl2021vision} to predict all Gaussian parameters. The depth head, \(\mathrm{F}_D\), ingests the image tokens \(\hat{t}_i^I\) and outputs per‑pixel depth maps \(D_i\) (with associated confidence \(C_i^D\)); these depths are then back‑projected through the predicted camera poses \(p_i\) to yield each Gaussian’s center \(\{\boldsymbol{\mu}_g\}_{g=1}^G\). The Gaussian head $F_G$ combines DPT features via \(\mathrm{F}_d(\hat{t}^I)\) with shallow CNN–extracted appearance features \(\mathrm{F}_a(I)\), and feeds their sum into a final regression CNN \(\mathrm{F}_b\) to predict opacity \(\sigma_g\), orientation \(\boldsymbol{r}_g\), scale \(\boldsymbol{s}_g\), SH color coefficients \(\boldsymbol{c}_g\), and per‑Gaussian confidence \(C_g\). Formally:
\begin{equation}
\begin{aligned}
    (D_i,\,C_i^D) &= \mathrm{F}_D(\hat{t}_i^I), \\
    \{\boldsymbol{\mu}_g\}     &= \mathrm{proj}\bigl(\{p_i\},\,\{D_i\}\bigr), \\
    \{\,\sigma_g,\boldsymbol{r}_g,\boldsymbol{s}_g,\boldsymbol{c}_g,C_g\} &= \mathrm{F}_b\bigl(\mathrm{F}_d^2(\{\hat{t}_i^I\}) + \mathrm{F}_a(\{I_i\})\bigr).
\end{aligned}
\end{equation}

\paragraph{Differentiable Voxelization}
\label{sec:diff_voxel}
Existing feed‑forward 3DGS methods typically assign one Gaussian per pixel, which works for sparse‑view inputs (2–16 images) but struggles with scaled-up complexity once more than 32 views are used. To address this, building upon \cite{scaffoldgs}, we introduce a differentiable voxelization module that clusters the \(G\) Gaussian centers \(\{\boldsymbol{\mu}_g\}\) into \(S\) voxels of size \(\epsilon\):

% \tl{give the value of $\epsilon$ in implementation details}

\begin{equation}
\{\boldsymbol{V}_s\}_{s=1}^S
\;=\;
\left\lfloor
\frac{\{\boldsymbol{\mu}_g\}_{g=1}^G}{\epsilon}
\right\rceil,
\end{equation}
where \(\boldsymbol{V}_s \in \{1,\dots,S\}\) denotes the voxel index of Gaussian \(g\).  

To keep voxelization differentiable, each Gaussian also predicts a confidence \(C_g\). We convert these scores into intra‑voxel weights via softmax:
\begin{equation}
w_{g\to s}
=\frac{\exp(C_g)}{\sum_{h:\boldsymbol{V}^h=s}\exp(C_h)}.
\end{equation}
% \textcolor{red}{TODO: check below please}
Finally, any per‑pixel Gaussian attribute \(a_g\) (e.g., opacity or color) is aggregated into its voxel by
\begin{equation}
\bar{a}_s
=\sum_{g:\boldsymbol{V}^g=s}w_{g\to s}\,a_g.
\end{equation}

The output of our pipeline is parameterized by the Gaussian attribute $\left\{ \left( \boldsymbol{\mu}_v, \sigma_v, \boldsymbol{r}_v, \boldsymbol{s}_v, \boldsymbol{c}_v \right) \right\}_s$ of each voxel \(\boldsymbol{V}_s \in \{1,\dots,S\}\). We can efficiently
render the Gaussians predicted by our model using differentiable Gaussian rasterization ~\cite{kerbl20233d, gsplat}.
This strategy dramatically reduces the number of primitives to process and enables end‑to‑end learning. 
% \tl{give an example ratio here?}
% }

\subsection{Training and Inference}
\label{sec:strategy}
\paragraph{Geometry Consistency Enhancement}
Predicting depth maps and camera poses simultaneously introduces subtle ambiguities that stem from multiview alignment and aggregation: when lifting per-image predictions to 3D, these inconsistencies manifest as layered sheets in the reconstructed point cloud, which may go unnoticed in raw point-cloud form but become glaringly obvious in rendered views. Such layering not only degrades visual fidelity but also prevents our outputs from meeting human-perceptual quality standards. To mitigate this, we introduce a geometry consistency loss that enforces agreement between rendered appearances and the underlying depth predictions, effectively smoothing out these layers and restoring coherent surface geometry.

Specifically, we enforce alignment between the depth maps \(D_i\) obtained from the DPT head \(\mathrm{F}_D\) and the rendered depth maps \(\hat{D}_i\) from 3D Gaussians. 
Since \(D_i\) can be unreliable in challenging regions, e.g., the sky or reflective surfaces, we utilize the jointly learned confidence map \(C^D_i\) and apply supervision only to the top $N \%$ of pixels with the confidence, ensuring that supervision focuses on the most trustworthy predictions. We align two depth maps as:
\begin{equation}
    \mathcal{L}_g = \frac{1}{N} \sum_{i=1}^{n} (D_i[M] - \hat{D}_i[M])^2,
    \label{eq:lg}
\end{equation}
where \(M\) is a geometry mask corresponding to the top $N$-quantile of the confidence map, we set $N = 30\%$ in our experiments.

Furthermore, we observed that, in the absence of supervision from novel views, the model tends to overfit to context views in an attempt to avoid interference from varying viewpoints. This results in poor generalization and leads to failures in depth and camera prediction. To mitigate this, we leverage a powerful pre-trained transformer network~\cite{wang2025vggt} to distill both camera parameters and scene geometry for stable training. Specifically, we regularize the camera parameters using the following loss function:
\begin{equation}
    \mathcal{L}_p = \frac{1}{N} \sum_{i=1}^N\left\|\tilde{p}_i-p_i\right\|_\epsilon,
    \label{eq:lp}
\end{equation}
where $\tilde{p}_i$ represents the pseudo ground-truth pose encoding, and $\left\|\cdot\right\|_\epsilon$ denotes the Huber loss. We then distill geometric information using:
\begin{equation}
    \mathcal{L}_d = \frac{1}{N} \sum_{i=1}^{n} (\tilde{D}_i[M] - \hat{D}_i[M])^2,
    \label{eq:ld}
\end{equation}
where \(\tilde{D}\) is the pseudo depth map obtained from the pre-trained model~\cite{wang2025vggt}. 
Experimental results show that this distillation loss significantly improves training stability and helps avoid convergence to poor local minima.

\paragraph{Training Objective} To avoid noises in the input data and better scale up data, \themethod is trained without any 3D supervision, using a pseudo-label training approach. Specifically, given a set of unposed and uncalibrated multi-view images \(\{I_i\}_{i=1}^N\) as input, our method first predicts their camera intrinsics and extrinsics. These predicted parameters are first used to project the positions of Gaussian primitives, and then rendered to produce the final outputs \(\{\hat{I}_i\}_{i=1}^N\).
Note that, although our model trains with only context views without novel views, \themethod presents great performance in novel view rendering due to the distill functions and great scene modeling capacity.

Finally, we optimize our model using a set of unposed images. We minimize the following loss function:

\begin{equation}
\begin{aligned}
    \mathcal{L} &= \mathcal{L}_{\text{rgb}} + \lambda_2 \cdot \mathcal{L}_g + \lambda_3 \cdot \mathcal{L}_p + \lambda_4 \cdot \mathcal{L}_d \\
    \mathcal{L}_{\text{rgb}} &= \operatorname{MSE}(I, \hat{I}) + \lambda_1 \cdot \operatorname{Perceptual}(I, \hat{I})
\end{aligned}
\end{equation}

\paragraph{Test-Time Camera Pose Alignment (Only for calculating the rendering metrics.)}
During inference, both the context views $\mathcal{I}_c$ and target views $\mathcal{I}_t$ are provided as inputs, where $\mathcal{I}_c \cap \mathcal{I}_t = \emptyset$. We assume the first frame of $\mathcal{I}_c$ is identical to the first frame of $\mathcal{I}_c \cup \mathcal{I}_t$. Consequently, the rotation of $\mathcal{I}_c$ and the context portion of $\mathcal{I}_c \cup \mathcal{I}_t$ remains the same; the only distinction lies in their scale. 
To address this, we compute the average context scale factor $s$ from $\mathcal{I}_c$ and the average scale factor $\hat{s}$ from $\mathcal{I}_c \cup \mathcal{I}_t$. The target scale is then normalized by multiplying it by the ratio $s / \hat{s}$.

% Next, we keep Gaussian parameters frozen, only refine the aligned target pose by optimizing the same rendering losses used for model training. Inspired by~\cite{monogs}, we compute the camera Jacobian matrix during optimization to reduce computational costs and accelerate the process.

\paragraph{Post Optimization (Optional)}
% We also provide a post-optimization option to further enhance reconstruction quality. 
We also include an optional post-optimization stage to further refine reconstructions, especially when inputs are dense.
After \themethod predicts the initial set of Gaussians and camera parameters, we first prune Gaussians with low opacity value (less than 0.01), and then render images from the input camera views and compute the MSE loss and the SSIM loss  between the rendered and input images.
We back-propagate the gradients through the Gaussian and camera parameters.  The learning rates are set as follows: 1.6e-4 for position, 5e-3 for scale, 1e-3 for rotation, 5e-2 for opacity, 2.5e-3 for color, and 5e-3 for camera pose.
\section{Experiments}
\subsection{Experimental Setup}
\paragraph{Datasets}
Following the common practice of CUT3R~\cite{wang2025continuous} and DUST3R~\cite{wang2024dust3r}, we train our model using images from nine public datasets: Hypersim~\cite{roberts2021hypersim}, ARKitScenes~\cite{baruch2021arkitscenes}, BlendedMVS~\cite{yao2020blendedmvs}, ScanNet++~\cite{yeshwanth2023scannet++}, CO3D-v2~\cite{reizenstein2021common}, Objaverse~\cite{deitke2023objaverse}, Unreal4K~\cite{tosi2021smd}, WildRGBD~\cite{xia2024rgbd}, and DL3DV~\cite{ling2024dl3dv}.
These datasets collectively span synthetic and real-world content, indoor and outdoor scenes, and object- to city-scale settings. This diverse data composition exposes the model to wide-ranging geometric and appearance variations, enhancing its generalization to unseen scenarios.

\begin{table*}[t!]
\caption{Quantitative Comparison on both sparse-view NVS setting (the number of input images is fewer than 16) and dense-view NVS setting (the number of input images is more than 32) on Mip-NeRF360~\cite{barron2022mip} and VR-NeRF~\cite{xu2023vr} dataset. We report both 3D scene reconstruction time and rendering quality metrics. We omit reporting the times for VR-NeRF, as its timings are consistent with the input values.}
\centering
\renewcommand{\arraystretch}{1.15}
\setlength{\tabcolsep}{2pt}
\resizebox{1\linewidth}{!}{
\begin{tabular}{l|cccc|cccc|cccc|l|cccc|cccc|cccc}
\toprule
\multicolumn{1}{c|}{\multirow{2}{*}{Sparse}} & \multicolumn{4}{c|}{3 Views} & \multicolumn{4}{c|}{6 Views} & \multicolumn{4}{c|}{16 Views} & \multicolumn{1}{c|}{\multirow{2}{*}{Dense}} & \multicolumn{4}{c|}{32 Views} & \multicolumn{4}{c|}{48 Views} & \multicolumn{4}{c}{64 Views}  \\
\cmidrule(lr){2-5} \cmidrule(lr){6-9} \cmidrule(lr){10-13} \cmidrule(lr){15-18} \cmidrule(lr){19-22} \cmidrule(lr){23-26}
 & PSNR$\uparrow$ & SSIM$\uparrow$ & LPIPS$\downarrow$ & Time(s)$\downarrow$ & PSNR$\uparrow$ & SSIM$\uparrow$ & LPIPS$\downarrow$ & Time(s)$\downarrow$ & PSNR$\uparrow$ & SSIM$\uparrow$ & LPIPS$\downarrow$ & Time(s)$\downarrow$ & & PSNR$\uparrow$ & SSIM$\uparrow$ & LPIPS$\downarrow$ & Time$\downarrow$ & PSNR$\uparrow$ & SSIM$\uparrow$ & LPIPS$\downarrow$ & Time$\downarrow$ & PSNR$\uparrow$ & SSIM$\uparrow$ & LPIPS$\downarrow$ & Time$\downarrow$ \\
\midrule
\multicolumn{26}{l}{\textit{Mip-NeRF360~\cite{barron2022mip} Dataset}} \\
NoPoSplat~\cite{ye2024no} & \textbf{16.36} & \underline{0.430} & \underline{0.453} & \textbf{0.119} & \underline{15.92} & \underline{0.416} & \underline{0.541} & \textbf{0.290} & \underline{15.47}  & \underline{0.361} & \underline{0.606} & \underline{1.198} &
3D-GS~\cite{kerbl20233d}      & \underline{22.19} & 0.640 & \underline{0.248} & \underline{10min} & \underline{21.86} & \underline{0.636} & \underline{0.274} & \underline{10min} & \underline{21.71} & \underline{0.626} & 0.300 & \underline{10min} \\
Flare~\cite{zhang2025flare}    &  13.52 & 0.350 & 0.601 & 0.271 & 15.35 & 0.407 & 0.573 & 0.415 & 13.21 & 0.348 & 0.695 & 1.201 &
Mip-Splatting~\cite{yu2024mip} & 22.07 & \underline{0.643} & 0.256 & 11min & 21.79 & 0.625 & 0.275 & 11min & \textbf{21.78} & \textbf{0.638} & \underline{0.299} & 11min \\
Ours       & \underline{16.20}  & \textbf{0.550} & \textbf{0.349} & \underline{0.171} & \textbf{18.32} & \textbf{0.524} & \textbf{0.329} & \underline{0.297} & \textbf{21.85} & \textbf{0.670} & \textbf{0.250} & \textbf{0.767} &
Ours       & \textbf{22.31} & \textbf{0.688} & \textbf{0.247} & \textbf{1.4s} & \textbf{21.90} & \textbf{0.652} & \textbf{0.273} & \textbf{2.7s} & 21.15 & 0.589 & \textbf{0.272} & \textbf{4.1s} \\

\cmidrule[0.5pt]{1-26}

\multicolumn{26}{l}{\textit{VR-NeRF~\cite{xu2023vr} Dataset}} \\
NoPoSplat~\cite{ye2024no} & 18.37 & 0.707 & 0.437 &   & 17.57 & 0.704 & \underline{0.466} &   & \underline{17.66} & \underline{0.720} & \underline{0.472} &  & 3D-GS~\cite{kerbl20233d}      & 22.37 & \underline{0.774} & \underline{0.302} &   & \textbf{22.86} & \underline{0.780} & \underline{0.306} &  & \underline{22.10} & \underline{0.770} & \underline{0.315} &  \\
Flare~\cite{zhang2025flare}     & \underline{18.58} & \underline{0.717} & \underline{0.470} & -- & \underline{18.26} & \underline{0.717} & 0.477 & -- & 17.02 & 0.709 & 0.510 & -- & Mip-Splatting~\cite{yu2024mip} & \underline{22.41} & 0.768 & 0.316 & -- & 22.55 & 0.772 & 0.314 & -- & 21.75 & 0.760 & 0.326 & -- \\
Ours      & \textbf{20.63} & \textbf{0.738} & \textbf{0.339} &  & \textbf{21.57} & \textbf{0.729} & \textbf{0.356} &  & \textbf{22.32} & \textbf{0.784} & \textbf{0.304} &  & Ours       & \textbf{23.09} & \textbf{0.781} & \textbf{0.230} &  & \underline{22.58} & \textbf{0.785} & \textbf{0.238} &  & \textbf{22.13} & \textbf{0.779} & \textbf{0.250} &  \\

% \cmidrule[0.5pt]{1-26}
% \multicolumn{26}{l}{\textit{Deep Blending~\cite{hedman2018deep} Dataset}} \\ 
% NoPoSplat~\cite{ye2024no} & 14.99 & 0.443 & 0.612 &  & 15.38 & 0.367 & 0.668 &  & 13.22 & 0.457 & 0.729 & & 3D-GS~\cite{kerbl20233d}      & 15.20 & \textbf{0.528} & \underline{0.490} &  & \underline{17.78} & 0.575 & \underline{0.400} &  & \textbf{17.11} & \textbf{0.557} & \underline{0.471} &   \\
% Flare~\cite{zhang2025flare}     & \underline{18.12} & \underline{0.562} & \underline{0.513} & -- & \underline{16.83} & \underline{0.477} & \underline{0.605} & -- & \underline{16.05} & \textbf{0.611} & \underline{0.617} & -- & Mip-Splatting~\cite{yu2024mip} & \underline{15.25} & \underline{0.521} & 0.504 & -- & 17.52 & \underline{0.586} & 0.451 & -- & 16.83 & 0.537 & 0.488 & --\\
% Ours      & \textbf{19.83} & \textbf{0.579} & \textbf{0.321} &  & \textbf{21.29} & \textbf{0.508} & \textbf{0.345} &  & \textbf{18.53} & \underline{0.524} & \textbf{0.435} &  & Ours       & \textbf{15.75} & 0.479 & \textbf{0.485} &  & \textbf{17.98} & \textbf{0.606} & \textbf{0.365} &  & \textbf{17.11} & \underline{0.553} & \textbf{0.425} &  \\

\bottomrule
\end{tabular}}
\label{tab:nvs}
\end{table*}

\paragraph{Training View Sampling Strategy}
View-sampling strategy is crucial for ensuring model robustness. We apply three different strategies depending on the dataset type. For object-centric datasets such as CO3D-v2~\cite{reizenstein2021common}, Objaverse~\cite{deitke2023objaverse}, and WildRGBD~\cite{xia2024rgbd}, we randomly sample views within a selected capture sequence. 
For sequential datasets like ARKitScenes~\cite{baruch2021arkitscenes} and DL3DV~\cite{ling2024dl3dv}, we first define minimum and maximum temporal gaps, then randomly select a value within this range to determine the interval between the first and last frames; views are then randomly sampled from within this interval. 
For unordered datasets like Hypersim~\cite{roberts2021hypersim}, BlendedMVS~\cite{yao2020blendedmvs}, ScanNet++~\cite{yeshwanth2023scannet++}, and Unreal4K~\cite{tosi2021smd}, we sample views based on pose distances. Specifically, we randomly choose a reference frame, compute the pose distance from all other frames to this reference, and sample views based on a predefined distance threshold.

\paragraph{Implementation details}
We set layer number $L=24$ for the Alternating-Attention Transformer and initialize the geometry transformer and depth DPT head with weights from VGGT~\cite{wang2025vggt}, while the remaining layers are initialized randomly. During training, we freeze the patch embedding weights. The model has approximately 886 million parameters in total. For differentiable voxelization, we set the voxel size $\epsilon$ to 0.002.

We train the model using the AdamW optimizer for 15K iterations. A cosine learning rate scheduler is employed, with a peak learning rate of 2e-4 and a warmup phase of 1K iterations. For layers initialized from VGGT, the learning rate is scaled by a factor of 0.1.
We train \themethod on 16 NVIDIA A800 GPUs for approximately two days. To save GPU memory and accelerate training, we use FlashAttention, bfloat16 precision, and gradient checkpointing. For stable training, we also skip optimization steps where the total loss exceeds 0.2 after the first 1K iterations.
In each iteration, we first select a training dataset at random, where each dataset is sampled according to a predefined weight (Fig.~\ref{tab:append_data_statics}). 
From the chosen dataset, we randomly sample between 2 and 24 frames, while maintaining a constant total of 24 frames per GPU.
The maximum input resolution is set to 448 pixels on the longer side. The aspect ratio is randomized between 0.5 and 1.0. Additionally, we apply intrinsic augmentation by randomly center-cropping each image to between 77\% and 100\% of its original size. Images are also augmented via random flipping.
For the training objective, we set $\lambda_1=0.05$, $\lambda_2$=0.1, $\lambda_3$=10.0, and $\lambda_4$=1.0.

\paragraph{Baselines}
We establish our sparse-view novel view synthesis baseline using previous state-of-the-art pose-free feed-forward methods, including Flare~\cite{zhang2025flare} and NoPoSplat~\cite{ye2024no}. For each evaluation dataset, we select three sparse-view subsets per scene; details of the view selection policy are provided in the appendix. Notably, prior methods require a post-optimization step during evaluation to align predicted camera poses with ground truth. However, we observe that this often fails—especially when there is limited overlap between training views—and can even degrade performance by overfitting to regions not visible during training. To ensure a fair comparison, we propose a more robust alignment strategy: we fix the first predicted camera as the identity and transform all other predicted rotations into this reference coordinate system. For translation alignment, we compute the median camera distance and estimate a relative scale factor to align the predicted and ground truth translations.
For our dense-view novel view synthesis baseline, we compare against 3D Gaussian Splatting~\cite{kerbl20233d} and Mip-Splatting~\cite{yu2024mip}, which both train on 30K iterations. We use 32, 48, and 64 views for training, and select 4, 6, and 8 views for evaluation, respectively. Training and testing views are jointly sampled based on camera distance. Since COLMAP~\cite{schoenberger2016sfm} is often unreliable under sparse-view conditions, we use VGGT to calibrate the input images and generate a point cloud for initialization.

\paragraph{Metrics}\ To evaluate the quality of novel view synthesis, we compute PSNR, SSIM~\cite{wang2004image}, and LPIPS~\cite{zhang2018unreasonable} between the predicted images and the ground truth. Additionally, to assess the accuracy of the predicted relative image poses, we use the AUC metric, which measures the area under the accuracy curve across various angular thresholds. In our evaluation, we set thresholds as 5, 10, 20, and 30. Furthermore, to evaluate multi-view geometric consistency, we report two widely used depth consistency metrics: the Absolute Mean Relative Error (AbsRel), defined as:
\begin{equation}
    \text{AbsRel} = \frac{1}{M} \sum_{i=1}^M \frac{|\hat{D}_i - D_i|}{D_i},
\end{equation}

and the $\delta_1$ accuracy, which measures the percentage of pixels where
\begin{equation}
    \max\left(\frac{\hat{D}_i}{D_i}, \frac{D_i}{\hat{D}_i}\right) < 1.25.
\end{equation}

\begin{figure*}[t!]
    \centering
    \includegraphics[width=\linewidth, trim=0 8 0 0, clip]{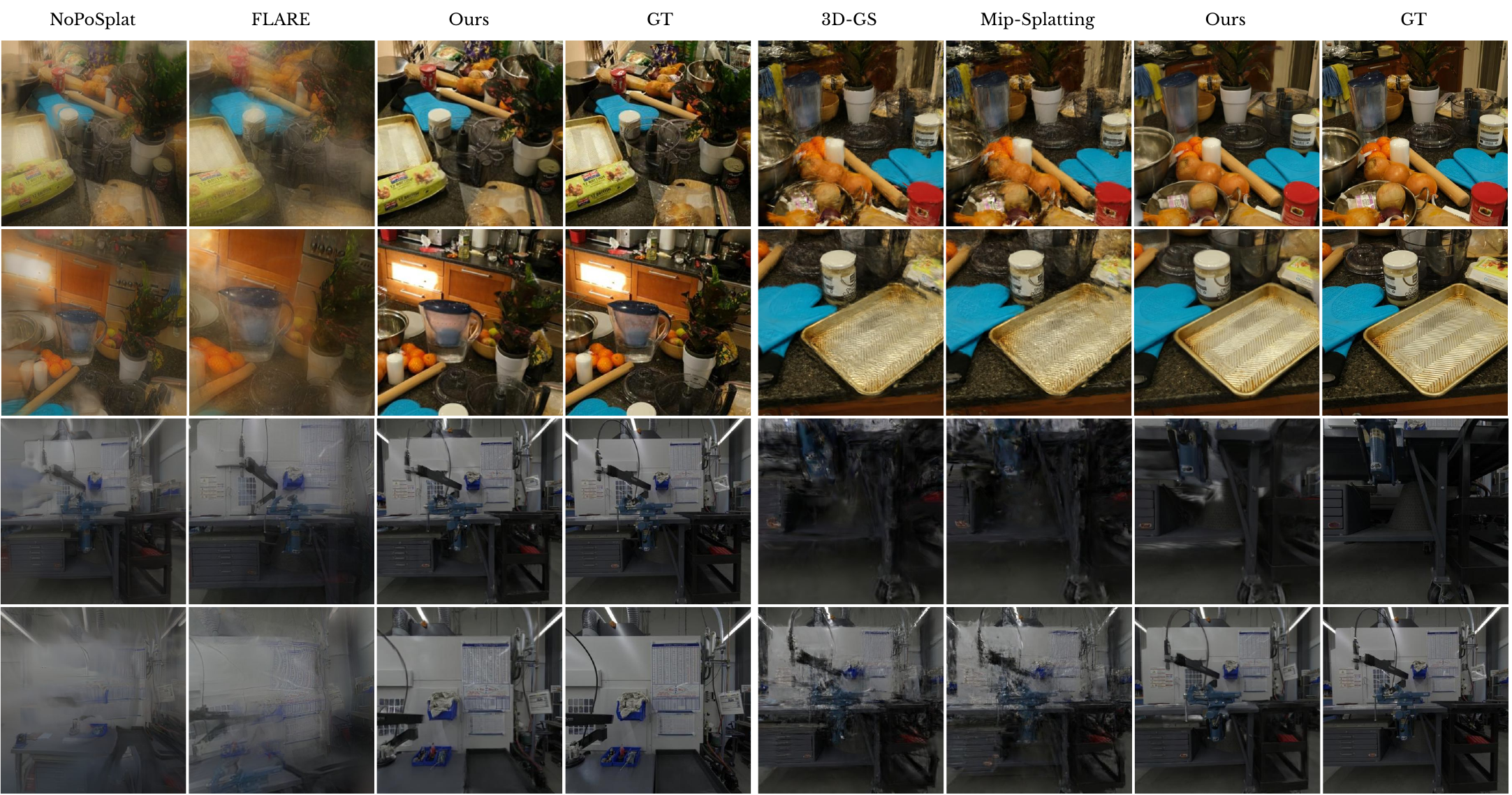}
    % \vspace{-1.5em}
    \caption{\label{fig:nvs_comparsion}%
        Visual comparisons between our method, NoPoSplat~\cite{ye2024no} and Flare~\cite{zhang2025flare} in the sparse view setting; 3D-GS~\cite{kerbl20233d} and Mip-Splatting~\cite{yu2024mip} in the dense view setting from two real-world datasets~\cite{barron2022mip,xu2023vr}. Our method shows excellent zero-shot performance, outperforming baselines in capturing sharp edges and intricate details.
    }
\end{figure*}

\subsection{Novel-view Synthesis}
Compared to prior pose-free feed-forward methods, which are typically limited to sparse-view inputs (e.g., 2–24 images), and optimization-based approaches that require up to 10 minutes per scene for dense-view reconstruction, our model generalizes to hundreds of input views and reconstructs 3D Gaussian primitives within just a few seconds on unseen scenes. We quantitatively evaluate our method against previous approaches on two zero-shot novel view synthesis datasets: MipNeRF-360~\cite{barron2022mip} and VR-NeRF~\cite{xu2023vr}, under both sparse-view and dense-view settings. 

\begin{figure*}[p]
    \centering
    \includegraphics[width=0.92\linewidth]{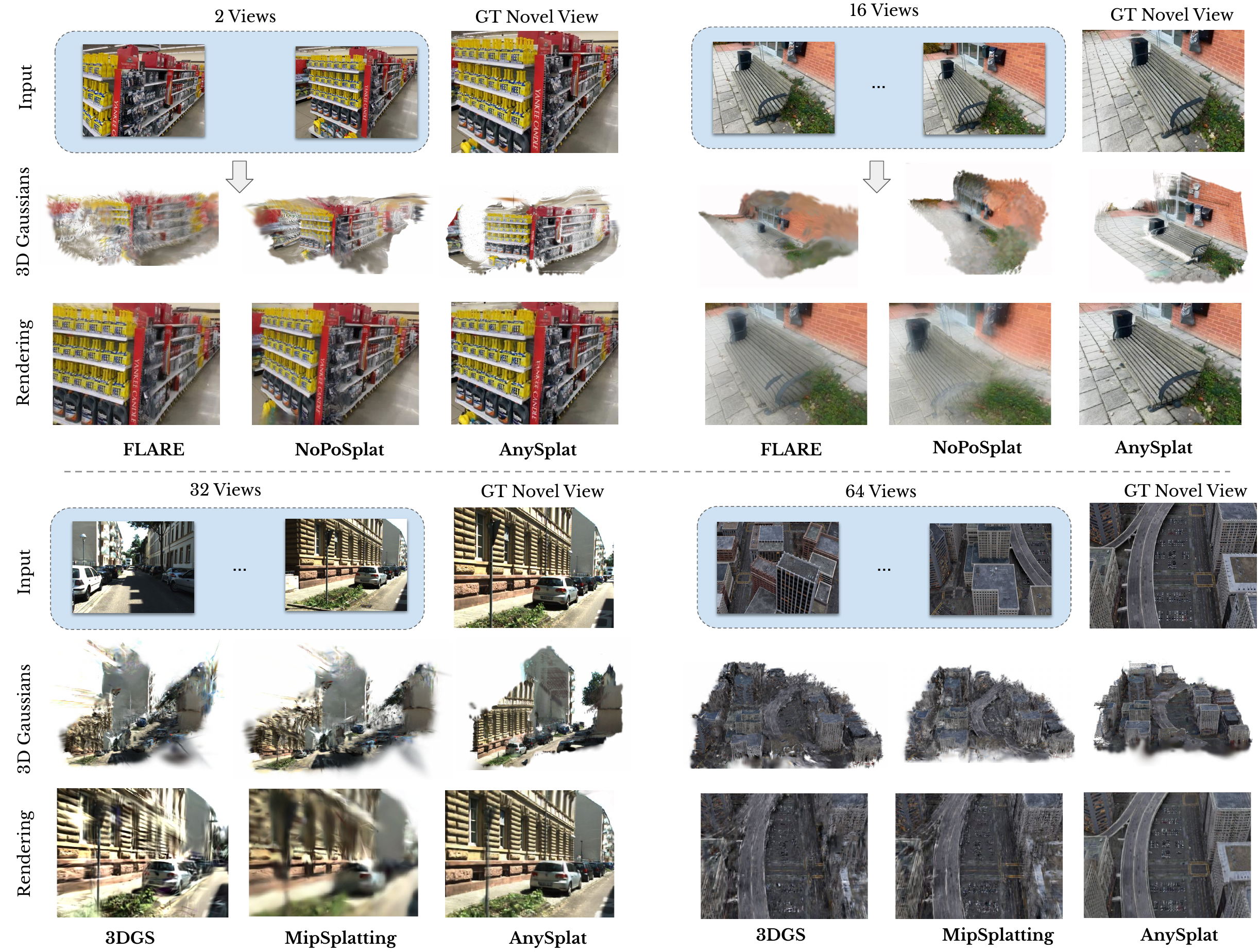}
    \vspace{-5pt}
    \caption{\label{fig:main-fig-page1}%
        \textbf{Qualitative comparisons} against baseline methods: for sparse-view inputs, we benchmark against the state-of-the-art FLARE~\cite{zhang2025flare} and NoPoSplat~\cite{ye2024no}; for dense-view inputs, we include 3DGS~\cite{kerbl20233d} and MipSplatting~\cite{yu2024mip} as representative comparisons.
        The slight misalignment between the rendered novel-views and the ground-truth is likely caused by pose-free reconstruction method’s estimated pose not perfectly matching the annotated ground-truth camera poses.
    }
\end{figure*}

% \boldparagraph{Sparse-view Novel-view Synthesis}
As shown in Tab.~\ref{tab:nvs}, Fig.~\ref{fig:nvs_comparsion} and Fig.~\ref{fig:main-fig-page1}, \themethod achieves improved rendering performance on sparse-view zero-shot datasets compared to recent feed-forward methods such as NoPoSplat~\cite{ye2024no} and Flare~\cite{zhang2025flare}.
There are two main reasons for this performance:
1) \themethod is trained on a diverse set of datasets and incorporates a random input view selection strategy, which contributes to its superior zero-shot generalization;
2) it achieves more accurate geometry and pose estimation, and since rendering quality strongly depends on pose accuracy, this leads to better visual results.
Moreover, with an increasing number of input views, our approach demonstrates faster inference times, which is important for real-world application.

In the dense-view setting (more than 32 views), \themethod continues to outperform optimization-based methods such as 3D-GS~\cite{kerbl20233d} and Mip-Splatting~\cite{yu2024mip} (with VGGT initialization), as shown in Tab.~\ref{tab:nvs}, Fig.~\ref{fig:nvs_comparsion} and Fig.~\ref{fig:main-fig-page1}. These optimization-based methods tend to overfit the training views, often resulting in artifacts in novel views. In contrast, our method reconstructs finer, cleaner geometry and delivers more detailed rendering results. Furthermore, \themethod achieves reconstruction times that are an order of magnitude faster than those of 3D-GS and Mip-Splatting.

\begin{figure}[t!]
  \centering
  \includegraphics[width=1.0\linewidth]{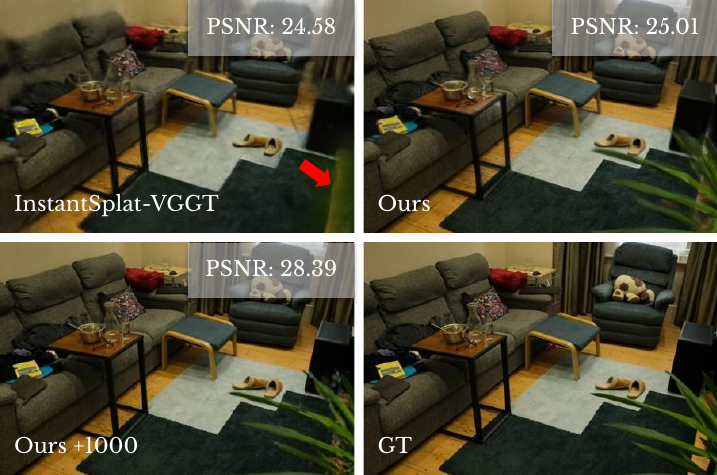}
  % \vspace{-6pt}
  \caption{Qualitative comparison of our method with InstantSplat-VGGT~\cite{fan2024instantsplat} style on Mip-NeRF360 dataset~\cite{barron2022mip} (room).}
  \label{fig:instantsplat}
  % \vspace{-1em}
\end{figure}

\begin{figure*}[p]
    \centering
    \includegraphics[width=0.85\linewidth]{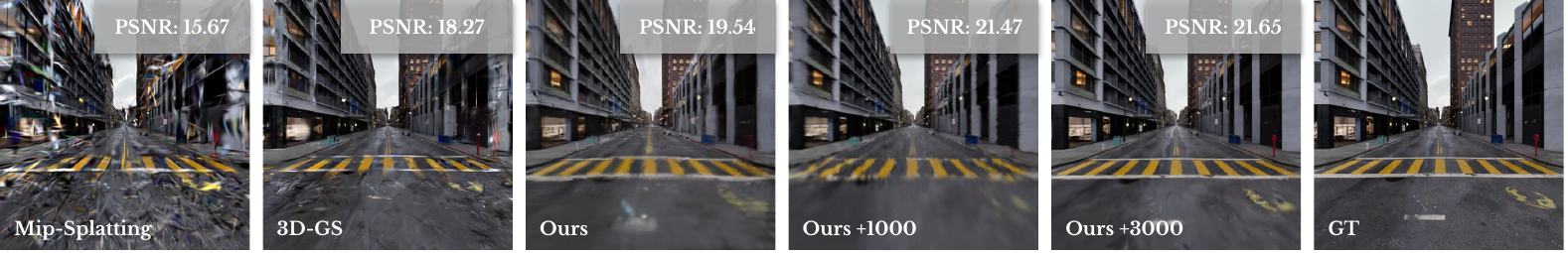}
    \vspace{-5pt}
    \caption{\label{fig:200views}%
        \textbf{Improved Rendering with Post-Optimization.} In our experiments using 200 input views, an optional post-optimization stage yields noticeably higher rendering fidelity, particularly in dense-view scenarios. 
    }
\end{figure*}

\paragraph{Post Optimization}
Although \themethod can efficiently perform end-to-end reconstruction of high-quality Gaussian models, further improvements can be achieved through an optional post-optimization step. 
As shown in Fig.~\ref{fig:200views} and Tab.~\ref{tab:200views}, we conduct a 200 input views experiment on the Matricity dataset~\cite{li2023matrixcity}. We demonstrate that even with 200 input, applying just 1000 steps of post-optimization (taking less than two minutes) yields improved results and 3000 steps can achieve much better results. Additionally, we conduct a 16-view experiment on the Mip-NeRF360 dataset~\cite{barron2022mip}, comparing our method with the InstantSplat-style~\cite{fan2024instantsplat} model, which is initialized using VGGT geometry predictions and optimized per scene with rendering losses over 1,000 iterations. As shown in Fig.~\ref{fig:instantsplat} and Tab.~\ref{tab:instantsplat}, our feed-forward approach achieves results comparable to InstantSplat-VGGT, and that post-optimization significantly improves performance.

\begin{table}[t!]
\caption{Quantitative comparison with 200 views on Matrixcity dataset~\cite{li2023matrixcity}. We compare our method, as well as its variants with 1K and 3K iters of post-optimization (\textit{Ours\_1000} and \textit{Ours\_3000}), against 3D-GS and Mip-Splatting.
}
% \vspace{-6pt}
\centering
\renewcommand{\arraystretch}{1.15}
\setlength{\tabcolsep}{4pt}
 \resizebox{0.95\linewidth}{!}{
\begin{tabular}{l|cccc}
\toprule 
Method  & PSNR$\uparrow$ & SSIM$\uparrow$ & LPIPS$\downarrow$ & Time $\downarrow$ \\
\midrule
3D-GS~\cite{kerbl20233d} & 19.10 & 0.614 & 0.450 & 10min \\
Mip-Splatting~\cite{yu2024mip} & 18.20 & 0.556 & 0.485 & 11min \\
% \midrule
Ours & 19.46 & 0.574 & \underline{0.446} & \textbf{33s} \\
Ours\_1000 & \underline{20.81} & \underline{0.635} & 0.519 & \underline{2min} \\
Ours\_3000 & \textbf{21.64} & \textbf{0.671} & \textbf{0.421} & 7min \\

\bottomrule
\end{tabular}}
\label{tab:200views}
\end{table}

\begin{table}[t!]
\caption{Quantitative comparison with 16 views on Mip-NeRF360 dataset~\cite{barron2022mip}. We compare our method, as well as 1K post-optimization (\textit{Ours\_1000}), against InstantSplat-VGGT~\cite{fan2024instantsplat} style.
}
% \vspace{-6pt}
\centering
\renewcommand{\arraystretch}{1.15}
\setlength{\tabcolsep}{4pt}
 \resizebox{0.95\linewidth}{!}{
\begin{tabular}{l|cccc}
\toprule 
Method  & PSNR$\uparrow$ & SSIM$\uparrow$ & LPIPS$\downarrow$ & Time $\downarrow$ \\
\midrule
InstantSplat-VGGT~\cite{fan2024instantsplat} & \underline{23.38} & \underline{0.677} & 0.268 & 3min \\
Ours & 21.85 & 0.670 & \underline{0.250} & \textbf{0.767s} \\
Ours\_1000 & \textbf{25.51} & \textbf{0.813} & \textbf{0.115} & \underline{2min} \\

\bottomrule
\end{tabular}}
\label{tab:instantsplat}
\end{table}

\subsection{Pose Estimation and Multi-view Geometry Consistency}
\themethod can be applied to relative pose estimation task. We evaluate it in a feed-forward setting on the RealEstate10K~\cite{zhou2018stereo} and CO3Dv2~\cite{reizenstein2021common} dataset with 10 randomly selected frames using a fixed seed for reproducibility, and compare its performance with VGGT, as shown in Tab.~\ref{tab:pose_esti}. Both two methods used Co3Dv2 samples in training, while RealEstate10K is excluded from the training set.
These results highlight the benefits of our rendering-based supervision, which slightly outperforms VGGT by enforcing stronger multi-view consistency constraints.

\begin{figure*}[p]
    \centering
    \includegraphics[width=0.86\linewidth]{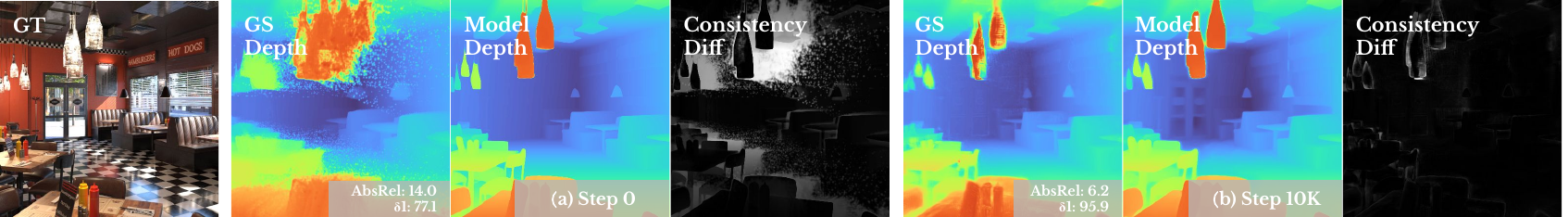}
    \vspace{-5pt}
    \caption{\label{fig:multiview-consis}%
        \textbf{Improvements of Multiview Consistency.} 
        From the initial iteration to 10k training steps, we observe a marked enhancement in multiview geometry consistency, clearly visible in the depth renderings, across both the model’s outputs and the 3D Gaussian Splatting renderings. This confirms the effectiveness of our geometry consistency enhancement design.
    }
\end{figure*}

In addition to pose estimation, we assess the multi-view geometric consistency of our approach. While VGGT~\cite{wang2025vggt} demonstrates strong performance in monocular depth prediction, it often suffers from poor consistency across views due to the lack of explicit 3D geometry constraints and its sensitivity to low-confidence regions, particularly around object boundaries. In contrast, our method leverages 3D rendering supervision to significantly enhance multi-view consistency. To evaluate this effect, we compare the depth maps rendered from Gaussians ($\hat{D}_i$) with those predicted by the DPT head ($D_i$) at both the beginning and end of training on the Hypersim dataset. As illustrated in Fig.~\ref{fig:multiview-consis}, the alignment between the two depth sources improves notably over training iterations, highlighting the effectiveness of our training strategy.

\begin{table}[t!]
\caption{Camera Pose Estimation on RealEstate10K~\cite{zhou2018stereo} and CO3Dv2~\cite{reizenstein2021common} with 10 random frames against VGGT~\cite{wang2025vggt}.}
% \vspace{-6pt}
\centering
\renewcommand{\arraystretch}{1.15}
\setlength{\tabcolsep}{2pt}
\resizebox{\linewidth}{!}{
\begin{tabular}{l|cccccccc}
\toprule 
\multicolumn{1}{c|}{\multirow{2}{*}{\ Method \ }} & \multicolumn{4}{c|}{RealEstate10K (unseen)} & \multicolumn{4}{c}{Co3Dv2}  \\
\cmidrule(lr){2-5} \cmidrule(lr){6-9} 
 & AUC@30$\uparrow$ & AUC@20$\uparrow$ & AUC@10$\uparrow$ & AUC@5$\uparrow$ & AUC@30$\uparrow$ & AUC@20$\uparrow$ & AUC@10$\uparrow$ & AUC@5$\uparrow$\\
\midrule
VGGT & 89.1 & 84.9 & 74.1 & 56.9 & 74.9 & 67.2 & 50.4 & 31.2\\
% \midrule
Ours & \textbf{89.2} & \textbf{85.1} & \textbf{74.6} & \textbf{57.9} & \textbf{78.3} & \textbf{71.6} & \textbf{56.9} & \textbf{39.2} \\

\bottomrule
\end{tabular}}
\label{tab:pose_esti}
\end{table}

\subsection{Ablation Study}
\label{sec:ab}
In this section, we ablate each individual module to validate their effectiveness. We conduct all the experiments based on the Hypersim Dataset. Quantitative and qualitative results can be found in Tab.~\ref{tab:ablation}. 

\begin{table}[t!]
\caption{Ablation Study. We evaluate the ablated variants of \themethod, discussed in Sec.~\ref{sec:ab}, by recording their rendering quality, geometric accuracy, and the size of the resulting Gaussian models.}
\centering
\renewcommand{\arraystretch}{1.15}
\setlength{\tabcolsep}{4pt}
\resizebox{1.0\linewidth}{!}{
\begin{tabular}{l|cccccc}
\toprule 
Method  & PSNR$\uparrow$ & SSIM$\uparrow$ & LPIPS$\downarrow$ & $\delta$1 $\uparrow$ & AbsRel$\downarrow$ & \#GS (M) \\
\midrule
Ours w/o Distill Loss & 7.28 & 0.217 & 0.832 & 75.5 & 14.7 &  4.80\\
Ours w/o Geo. Loss & \underline{18.20} & \underline{0.635} & \underline{0.285} & 94.7 & 7.6 & 3.52 \\
Ours w/o Diff. Voxel & 17.77 & 0.609 & 0.303 & 95.8 & \underline{5.7} & 4.82  \\
Ours frozen AA transformer layers & 17.90 & 0.616 & 0.306 & \textbf{96.5} & \textbf{5.3} & 3.51 \\
Ours frozen all transformer layers & 17.84 & 0.621 & 0.330 & 95.3 & 6.6 & \textbf{3.40} \\
% Ours w/ Linear Depth Head & 17.51 & 0.592 & 0.382 & 94.6 & 6.7 & 3.36 \\
% \midrule
Ours & \textbf{18.25} & \textbf{0.648} & \textbf{0.279} & \underline{96.3} & 5.9 & \underline{3.45}\\
\bottomrule
\end{tabular}}
% \vspace{-10pt}
\label{tab:ablation}
\end{table}

\paragraph{Distill Loss}
To evaluate the impact of the distillation losses defined in Eq.~\ref{eq:lp} and~\ref{eq:ld}, we perform an ablation study by removing them from training. As shown in Table~\ref{tab:ablation}, this leads to a significant drop in both rendering quality and geometric consistency. The results suggest that, in the absence of external supervision and when trained solely on unposed images, the model tends to overfit the input views with plausible renderings without preserving accurate 3D geometry. In our experiments, the absence of a distillation loss results in incorrect depth and pose predictions, leading to degraded performance in novel view renderings. The distillation loss mitigates this by reinforcing geometric consistency during training.

\paragraph{Geometry Consistency Loss}
We further demonstrate the effectiveness of our geometry consistency loss, defined in Eq.~\ref{eq:lg}, by comparing it against a variant of our model trained with only the rendering and distillation losses. As shown in the second and last rows of Table~\ref{tab:ablation}, incorporating the consistency loss encourages the model to produce more coherent multi-view geometry, resulting in a 1.7\% reduction in AbsRel and a 1.6\% improvement in $\delta_1$ accuracy.

\begin{figure}[t!]
    \centering
    \includegraphics[width=1\linewidth,trim=0 6 0 0,clip]{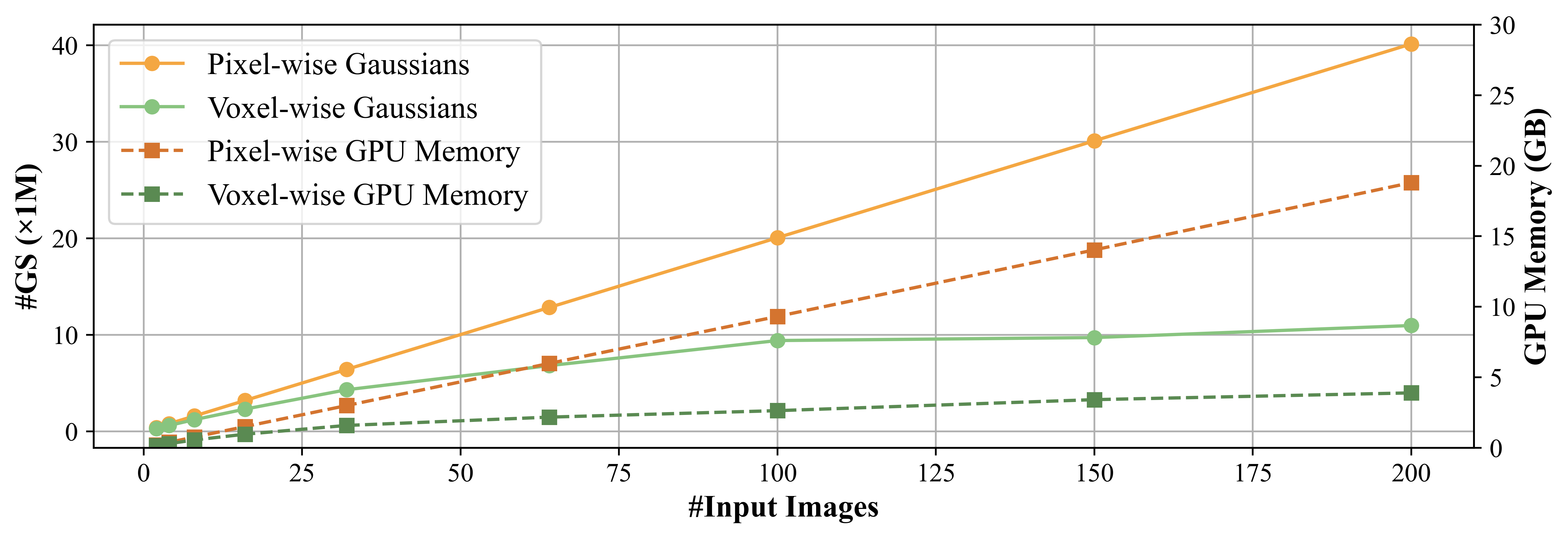}
    % \vspace{-2em}
    \caption{
    Growth of Gaussian Primitives and GPU Memory Usage.
    As the number of input views increases, the count of Gaussian primitives grows sublinearly and eventually plateaus when using the differentiable voxelization module. In contrast, without this module, the number of Gaussians increases approximately linearly. The GPU memory consumption for rendering mirrors this saturation behavior.
    }
    \label{fig:voxel_ratio}
\end{figure}

\paragraph{Differentiable Voxelization}
To evaluate the impact of the differentiable voxelization module introduced in Sec.~\ref{sec:pipeline}, we conduct an experiment in which this component is removed. Interestingly, the model achieves slightly better performance despite using fewer Gaussian primitives. This improvement can be attributed to the voxelization module’s ability to reduce redundancy among Gaussians and alleviate artifacts caused by overlapping primitives. 
Furthermore, as illustrated in Fig.~\ref{fig:voxel_ratio}, when differentiable voxelization is used, the number of Gaussians increases more slowly with the number of context views and eventually reaches saturation. This leads to lower GPU memory consumption during rendering compared to pixel-wise rendering approaches.

\paragraph{Training strategy}
We investigate different training strategies by exploring the following three experimental configurations:
\begin{itemize}
\item[1)] \textit{Frozen All Transformer:} All transformer layers initialized from VGGT are frozen during training, while the remaining parameters are trainable.
\item[2)] \textit{Frozen AA Transformer:} Only the Alternating-Attention layers are frozen, while the vision tokenizer is fine-tuned.
\item[3)] \textit{Frozen Vision Tokenizer:} The vision tokenizer is frozen, and only the Alternating-Attention layers are fine-tuned.
\end{itemize}
Our empirical results show that the third configuration yields the best performance, achieving PSNR gains of 0.41\,dB and 0.35\,dB over configuration 1 and 2, respectively. These findings suggest that preserving pre-trained visual representations while adapting the attention mechanism provides an effective balance between stability and adaptability during training.

\section{Conclusion and Future Works}

In this work, we introduce AnySplat, a feed-forward 3D reconstruction model that integrates a lightweight rendering head with our geometry-consistency enhancement, augmented by a pseudo-label knowledge distillation training strategy. We view this as a novel way to fully \emph{unlock} the potential of 3D foundation models and elevate their scalability to a broader scope.
Our experiments demonstrate AnySplat's robust and competitive results on both sparse and dense multiview reconstruction and rendering benchmarks using unconstrained, uncalibrated inputs. 
Additionally, the model training remains efficient, requiring minimal time and compute, enabling feed-forward 3D Gaussian Splatting reconstructions and high-fidelity renderings in just seconds at inference time. 
We expect this low-latency pipeline to open new possibilities for future interactive and real-time 3D applications.

Despite its improvements, AnySplat still observes artifacts in challenging regions, such as skies, specular highlights, and thin structures; its reconstruction-based rendering loss may be less stable under dynamic scenes or varying illumination, and the compute–resolution trade-off (i.e., number of Gaussians scaling alongside input and voxel resolution) can slow performance when handling very high resolution or large numbers of views. 
We see enhancing patch size flexibility, improving robustness to repetitive texture patterns, and streamlining scaling to thousands of high-resolution inputs as promising directions for future work.

\begin{acks}
This work was funded in part by the National Key R\&D Program of China (2022ZD0160201), Shanghai Artificial Intelligence Laboratory, the HKU Startup Fund, the HKU Shanghai Intelligent Computing Research Center, and the Anhui Provincial Natural Science Foundation under Grant 2108085UD12.
\end{acks}

% \clearpage
\bibliographystyle{ACM-Reference-Format}
\bibliography{sample-base}

% !TEX root = ./sample-acmtog.tex

%% Full-page figures following the main paper
% \input{tables/our-results-and-ablation.tex}

\begin{figure*}[p]
    \centering
    \includegraphics[width=\linewidth]{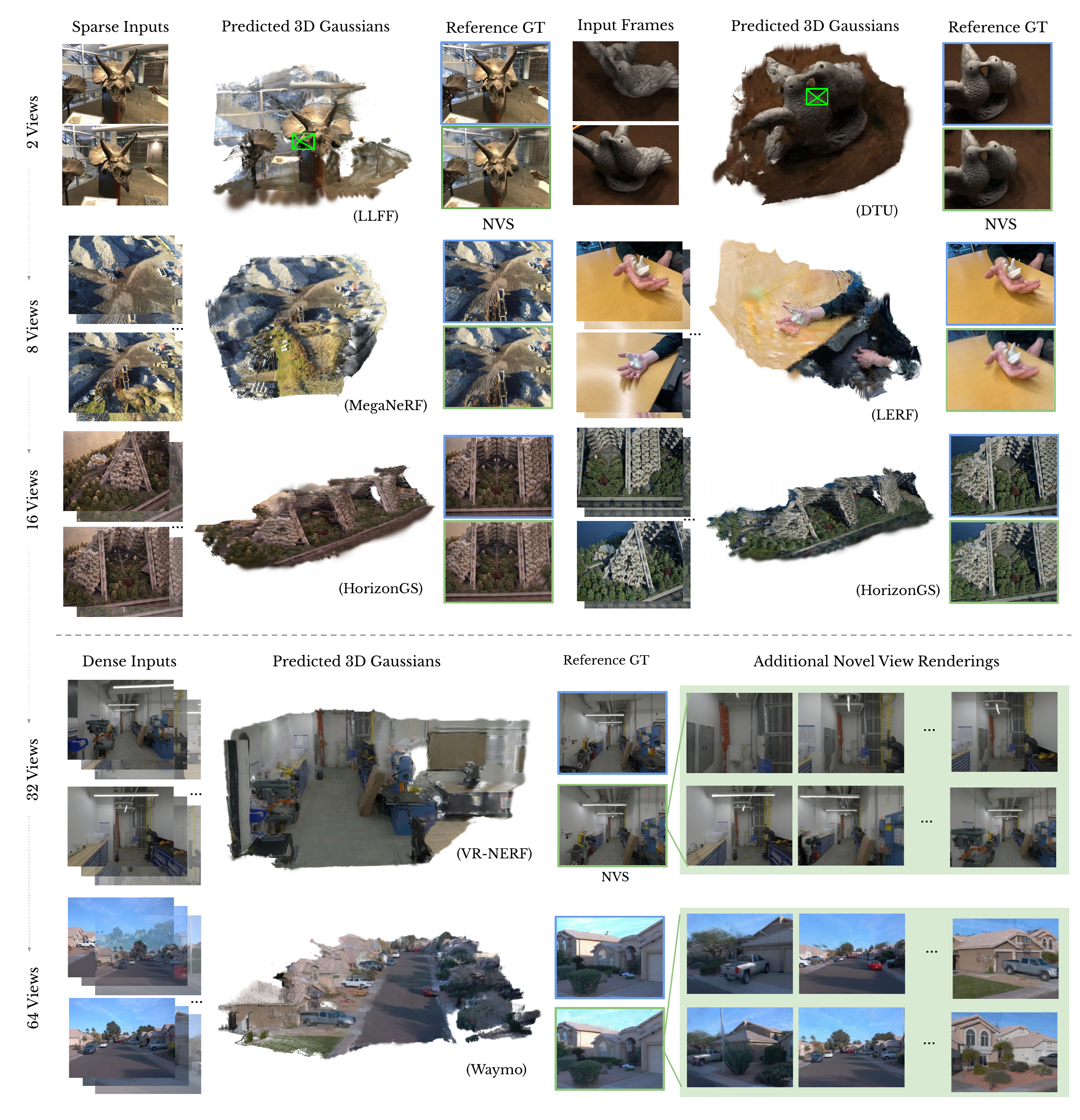}
    \caption{\label{fig:main-fig-page2}%
        % \textbf{Qualitative comparisons} 
        Example visualization of our AnySplat reconstruction and novel-view synthesis across a spectrum of scene complexities and input frames densities. 
        From top to bottom, the number of input images increases—from extremely sparse to medium and dense captures, while the scene scale grows from object-centric setups (LLFF~\cite{mildenhall2019local}, DTU~\cite{jensen2014large}) through mid-scale trajectories (MegaNeRF~\cite{turki2022mega}, LERF~\cite{kerr2023lerf}, HorizonGS~\cite{jiang2024horizon}) to large-scale indoor and outdoor environments (VR-NeRF~\cite{xu2023vr}, Waymo~\cite{sun2020scalability}). For each setting, we display the input views, the reconstructed 3D Gaussians, the corresponding ground-truth renderings, and example novel-view renderings.
    }
\end{figure*}

\appendix
\clearpage

The following appendices provide additional technical details and experimental results that support the main findings of this work. 
% For additional qualitative examples on further datasets, please see the accompanying demo video. 

\section{Experiment Details}
\label{sec:supp_expriment}

In this section, we provide additional details of our training protocol, model initialization, and experiments.

\paragraph{Training Setting}

We train on a heterogeneous mix of nine datasets spanning synthetic indoor scenes (Hypersim~\cite{roberts2021hypersim}, ARKitScenes~\cite{baruch2021arkitscenes}, BlendedMVS~\cite{yao2020blendedmvs}, ScanNet++~\cite{yeshwanth2023scannet++}, CO3D-v2~\cite{reizenstein2021common}, Objaverse~\cite{deitke2023objaverse}, Unreal4K~\cite{tosi2021smd}, WildRGBD~\cite{xia2024rgbd}, and DL3DV~\cite{ling2024dl3dv}). 
During each iteration we randomly sample one dataset according to the distribution shown in Tab.~\ref{tab:append_data_statics}, ensuring balanced exposure to both synthetic and real environments. This mixture stabilizes convergence and improves generalization across diverse scene types.
In future work, we plan to incorporate additional high-fidelity datasets, particularly from open-world simulators such as game engines that provide accurate 3D geometry and scene consistency—to further empower our model’s scalability. We also intend to include a wider variety of camera trajectories.

\begin{table}[h!]
\caption{
Training Datasets Statistics. 
% We randomly sample the dataset in each training iteration according to the probability shown in this table. \textit{Prob.} represent the sampling ptobabilty.
We report the sampling distribution over our nine training datasets: at each iteration, we randomly select one dataset according to the probabilities listed in the  Prob column, which reflects the relative frequency with which each dataset is drawn during training.
}
\centering
\renewcommand{\arraystretch}{1.15}
\setlength{\tabcolsep}{4pt}
\resizebox{\linewidth}{!}{
\begin{tabular}{l|ccccc}
\toprule 
Dataset  & Scene Type & Real/Synthetic & \# of Frames& \# of Scenes & Training Prob. (\%)  \\
\midrule
ARKitScenes~\cite{baruch2021arkitscenes} & Indoor & Real & 9.2M & 4406 & 16.7\\ 
ScanNet++~\cite{yeshwanth2023scannet++} & Indoor & Real & 1.0M & 935 & 16.7\\
BlendedMVS~\cite{yao2020blendedmvs} & Mixed & Real & 114K & 467 & 8.3\\
Unreal4K~\cite{tosi2021smd} & Mixed & Synthetic & 16K & 18 & 8.3\\
CO3Dv2~\cite{reizenstein2021common} & Object & Real & 5.5M & 27520 & 8.3\\ 
DL3DV~\cite{ling2024dl3dv} & Mixed & Real & 3.4M & 9894 & 16.7\\
WildRGBD~\cite{xia2024rgbd} & Object & Real & 3.9M & 11050 & 8.3\\
Hypersim~\cite{roberts2021hypersim} & Indoor & Synthetic & 73K & 744 & 8.3\\ 
Objaverse~\cite{deitke2023objaverse} & Object & Synthetic & 8M & 199K & 8.3\\
\bottomrule
\end{tabular}}
\label{tab:append_data_statics}
\end{table}

\paragraph{Model Initialization}
To leverage prior geometric structure, we initialize our geometry-transformer backbone with weights pretrained on the VGGT dataset. All parameters in the Gaussian-prediction head are drawn from a zero-mean Gaussian distribution with standard deviation 0.02, while all biases are set to zero. This strategy allows the geometry branch to start from a strong prior, accelerating convergence, while the Gaussian head learns scene appearance and density from scratch.

\paragraph{Evaluation Setting}
\label{sec:evaluation_setting}

We evaluate our approach on two widely used benchmarks: the VR-NeRF dataset~\cite{xu2023vr} and the Mip-NeRF360 dataset ~\cite{barron2022mip}. From VR-NeRF, which offers richly textured indoor environments with varied layouts, we randomly select four representative scenes—apartment, kitchen, raf-furnishedroom, and workshop—ensuring a mix of both compact and spacious rooms. From Mip-NeRF360, a dataset known for its challenging viewpoint diversity and complex lighting, we include all available scenes: bonsai, counter, kitchen and room. Together, these seven scenes cover a broad spectrum of indoor settings, camera densities, and appearance variations, allowing us to stress-test both sparse- and dense-view reconstruction scenarios.

In the dense-view setting, we select one out of every eight images as the test view. We first choose 72 images from the dataset, either randomly or based on spatial distribution. From these 72 images, we further sample subsets of 54 and 36 images. After excluding the test views, the numbers of input images for these three cases are 64, 48, and 32, respectively. In the sparse-view setting, we select one out of every two images as the test view. The view-selection procedure is the same as in the dense-view setting.

\begin{figure*}[b!]
    \centering
    \includegraphics[width=1\linewidth]{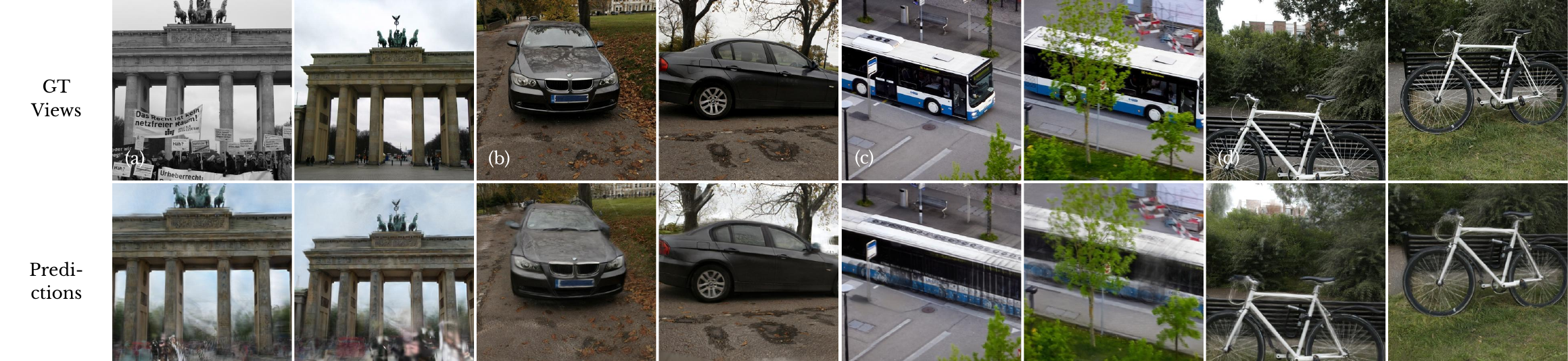}
    \caption{\label{fig:fail_case}%
        Example failure cases. AnySplat exhibits visible artifacts under (a) variable illumination or transient occluders for the Brandenburg Gate (Phototourism~\cite{jin2021image}); (b) specular highlights on the sedan (Ref-NeRF~\cite{verbin2022ref}); (c) a dynamic bus scene (DAVIS~\cite{perazzi2016benchmark}); and (d) the bicycle’s thin structures (Mip-NeRF360~\cite{barron2022mip}).
    }
\end{figure*}

\section{More Results}
\label{sec:more_results}

\paragraph{Same Test Views.} 
In Table~\ref{tab:nvs}, we use the view-selection strategy described in sec.~\ref{sec:evaluation_setting}. However, those results do not isolate how rendering performance depends on the number of input views. To make this dependence explicit, we compare 3D-GS~\cite{kerbl20233d} and Mip-Splatting~\cite{yu2024mip} using the same fixed test views in the dense-view setting (Table~\ref{tab:nvs_same}). Specifically, we sample 72 images and hold out 8 as test views; keeping these test views fixed, we then construct training sets of 64, 48, and 32 input images by randomly selecting from the remaining 64 images. This setup not only reveals the relationship between input count and performance but also rigorously evaluates our model’s sensitivity to view sampling and its robustness across arbitrary camera configurations.
These results lead to two conclusions: (1) in 3D scene reconstruction, more input views yield higher rendering quality for both feed-forward and per-scene optimization methods; and (2) AnySplat’s rendering quality is consistently competitive with per-scene optimization methods, underscoring the promise of feed-forward approaches.

\begin{table}[htbp]
\caption{Quantitative Comparison on dense-view NVS setting on Mip-NeRF360~\cite{barron2022mip} dataset with same test view images.}
\centering
\renewcommand{\arraystretch}{1.15}
\setlength{\tabcolsep}{2pt}
\resizebox{1\linewidth}{!}{
\begin{tabular}{l|ccc|ccc|ccc}
\toprule
\multicolumn{1}{c|}{\multirow{2}{*}{Method}} & \multicolumn{3}{c|}{32 Views} & \multicolumn{3}{c|}{48 Views} & \multicolumn{3}{c}{64 Views}  \\
% \cmidrule(lr){2-5} \cmidrule(lr){6-9} \cmidrule(lr){10-13} 
 & PSNR$\uparrow$ & SSIM$\uparrow$ & LPIPS$\downarrow$ & PSNR$\uparrow$ & SSIM$\uparrow$ & LPIPS$\downarrow$ & PSNR$\uparrow$ & SSIM$\uparrow$ & LPIPS$\downarrow$ \\
\midrule
3D-GS~\cite{kerbl20233d} & \textbf{17.25} & \underline{0.416} & \underline{0.439} & \underline{18.62} & \underline{0.461} & \underline{0.415} & \underline{19.05} & \underline{0.490} & \underline{0.417} \\
Mip-Splatting~\cite{yu2024mip} & \underline{16.95} & 0.381 & 0.450 & 18.39 & 0.457 & 0.424 & 18.94 & 0.486 & 0.427\\
Ours & 16.59 & \textbf{0.417} & \textbf{0.422} & \textbf{18.72} & \textbf{0.479} & \textbf{0.370} & \textbf{19.57} & \textbf{0.508} & \textbf{0.356}       \\

\bottomrule
\end{tabular}}
\label{tab:nvs_same}
\end{table}

\paragraph{Failure Case.}
Although AnySplat performs well on most scenes, we still observe some failure cases (Fig.~\ref{fig:fail_case}). For example, AnySplat can struggle with (a) variable illumination and transient occlusions, (b) specular highlights, (c) dynamic scenes, and (d) fine-grained geometry. The first three issues arise because these factors are not explicitly modeled; incorporating appropriate modeling strategies and richer training data could mitigate them. The last issue likely requires a more powerful geometry encoder. We leave these directions to future work.

\begin{figure*}[p]
    \centering
    \includegraphics[width=\linewidth]{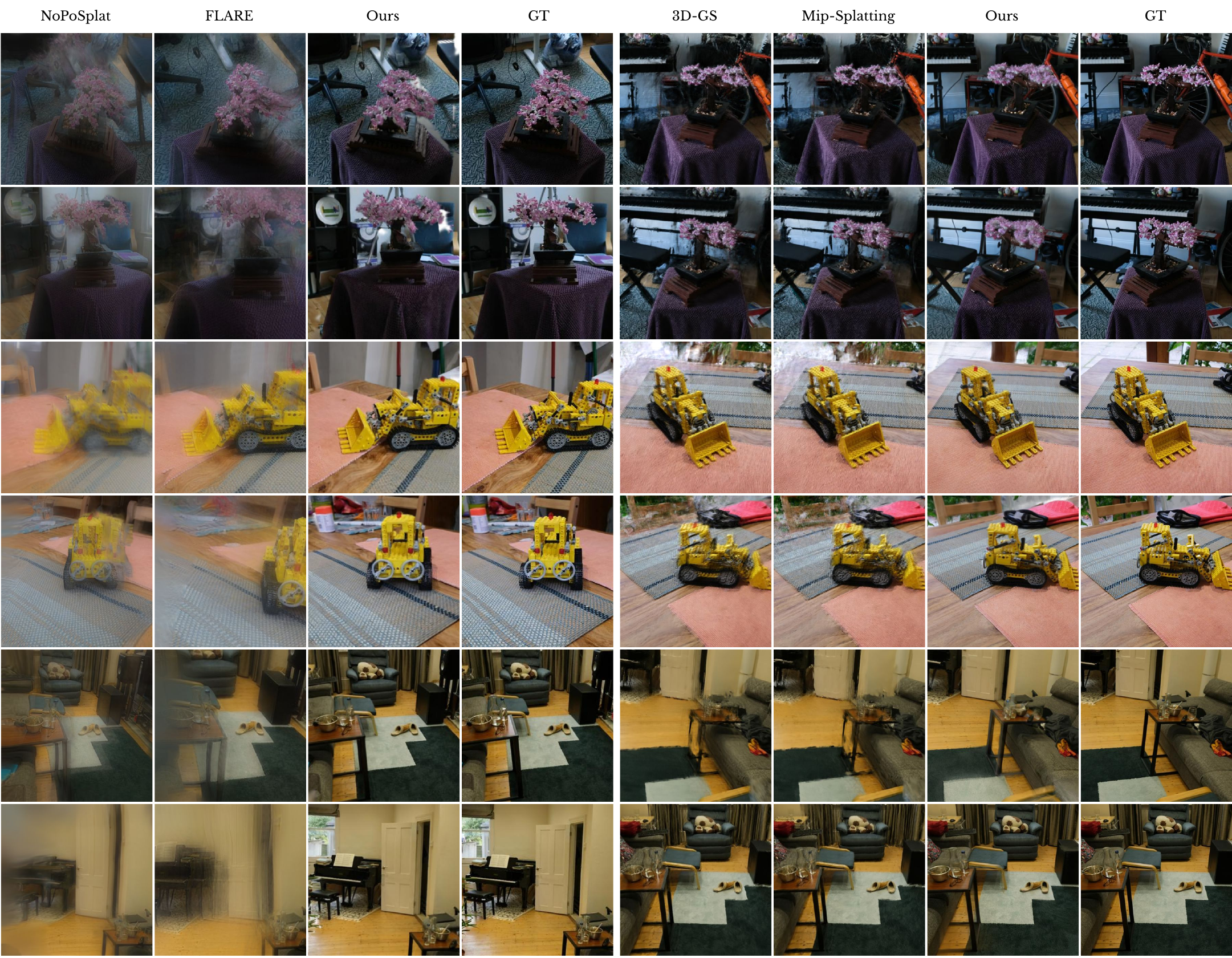}
    \caption{\label{fig:mipnerf360}%
        Example visualization results on the Mip-NeRF360~\cite{barron2022mip} dataset (bonsai, kitchen, room).
    }
\end{figure*}

\begin{figure*}[p]
    \centering
    \includegraphics[width=\linewidth]{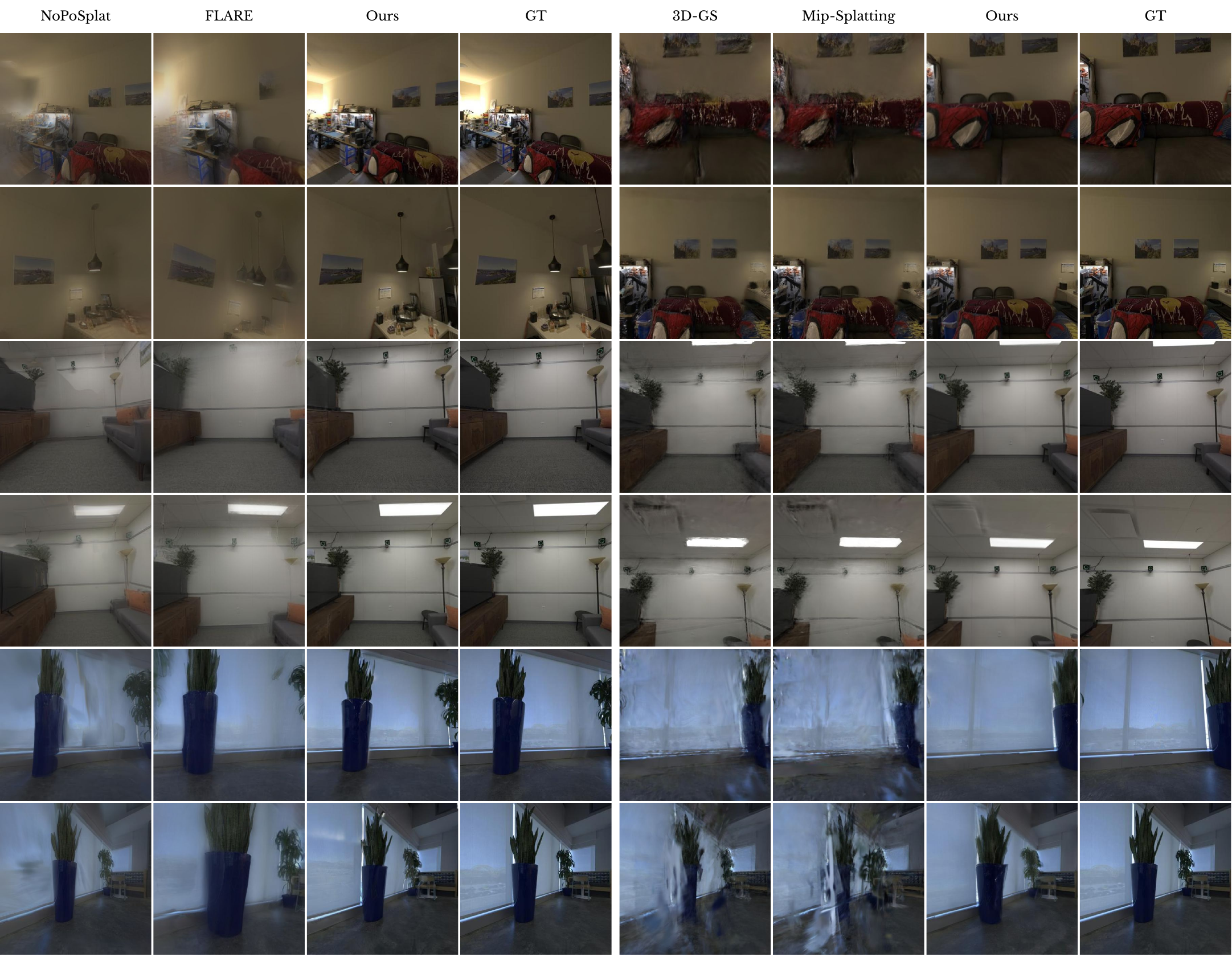}
    \caption{\label{fig:VR-NeRF}%
        Example visualization results on the VR-NeRF~\cite{xu2023vr} dataset (apartment, raf\_furnishedroom, kitchen).
    }
\end{figure*}

\paragraph{More Comparisons.}
We present more visualization results in Fig.~\ref{fig:mipnerf360} and Fig.~\ref{fig:VR-NeRF}. For sparse-view inputs, AnySplat delivers higher visual quality, with reliable geometry and finer details, than NoPoSplat~\cite{ye2024no} and Flare~\cite{zhang2025flare}. For dense-view inputs, 3D-GS~\cite{kerbl20233d} and Mip-Splatting~\cite{yu2024mip} tend to overfit in the training views, leading to unavoidable artifacts. In contrast, AnySplat consistently produces cleaner renderings with fewer artifacts.

\end{document}